\documentclass{article}

\usepackage[preprint]{neurips_2026}

\usepackage[utf8]{inputenc}
\usepackage[T1]{fontenc}
\usepackage{hyperref}
\usepackage{url}
\usepackage{graphicx}
\usepackage{subcaption}
\usepackage{wrapfig}
\usepackage{float}
\usepackage{booktabs}
\usepackage{multirow}
\usepackage{amsmath}
\usepackage{amssymb}
\usepackage{mathtools}
\usepackage{amsthm}
\usepackage{array}
\usepackage{arydshln}
\usepackage{makecell}
\usepackage{bm}
\usepackage{xcolor}
\usepackage{microtype}
\usepackage{pdfrender}
\usepackage[normalem]{ulem}
\usepackage[capitalize,noabbrev]{cleveref}
\newcommand{\TabMainResults}{
\begin{table*}[t]
\centering
\small
\renewcommand{\arraystretch}{1.06}
\addtolength{\tabcolsep}{-1.5pt}
\caption{
RQ1: Long-Horizon sequential model editing results. Reported values are mean$\pm$std for each metric. For each model: Pre-edited is separated by a rule; Non-LE baselines appear above the next rule; LE-family methods appear below. Best in \best{bold}, second-best in \second{underlined}.
}
\label{tab:seq_edit_long}

\resizebox{\textwidth}{!}{%
\begin{tabular}{@{}c l n n n n n n n n@{}}
\toprule
\multirow{2}{*}{\textbf{Model}} & \multirow{2}{*}{\textbf{Method}}
& \multicolumn{5}{c@{}}{\textbf{CounterFact}}
& \multicolumn{3}{@{}c}{\textbf{ZsRE}} \\
\cmidrule(r{10pt}){3-7}\cmidrule(l{10pt}){8-10}
& & \textbf{Eff.$\uparrow$} & \textbf{Gen.$\uparrow$} & \textbf{Spe.$\uparrow$}
  & \textbf{Flu.$\uparrow$} & \textbf{Consis.$\uparrow$}
  & \textbf{Eff.$\uparrow$} & \textbf{Gen.$\uparrow$} & \textbf{Spe.$\uparrow$} \\
\midrule

\multirow{10}{*}{\vmodel{LLaMA3}}%
& Pre-edited      & \pmstd{7.18}{0.13} & \pmstd{9.39}{0.09} & \pmstd{89.79}{0.21} & \pmstd{568.53}{1.11} & \pmstd{3.55}{0.04} & \pmstd{36.11}{0.24} & \pmstd{35.06}{0.37} & \pmstd{32.09}{0.30} \\
\presep
& FT                & \second{\pmstd{68.69}{0.30}} & \second{\pmstd{61.99}{0.21}} & \pmstd{40.91}{0.38} & \pmstd{325.90}{0.41} & \pmstd{1.44}{0.05} & \pmstd{8.58}{0.04} & \pmstd{6.96}{0.05} & \pmstd{17.30}{0.23} \\
& UltraEdit      & \pmstd{60.06}{0.20} & \pmstd{52.90}{0.18} & \pmstd{43.88}{0.14} & \pmstd{558.79}{0.70} & \pmstd{11.70}{0.34} & \pmstd{77.61}{0.13} & \pmstd{75.15}{0.33} & \best{\pmstd{50.26}{0.13}} \\
& RLEdit           & \pmstd{65.38}{0.29} & \pmstd{49.18}{0.26} & \pmstd{47.86}{0.30} & \second{\pmstd{596.19}{0.11}} & \second{\pmstd{12.18}{0.33}} & \second{\pmstd{81.59}{0.35}} & \second{\pmstd{78.77}{0.17}} & \pmstd{27.03}{0.35} \\
\blocksep
& MEMIT           & \pmstd{50.95}{0.14} & \pmstd{50.95}{0.19} & \pmstd{49.05}{0.23} & \pmstd{244.21}{0.29} & \pmstd{4.86}{0.13} & \pmstd{0.95}{0.03} & \pmstd{0.22}{0.06} & \pmstd{0.22}{0.06} \\
& PRUNE           & \pmstd{49.88}{0.21} & \pmstd{49.87}{0.13} & \second{\pmstd{50.13}{0.27}} & \pmstd{495.37}{0.18} & \pmstd{2.77}{0.11} & \pmstd{0.00}{0.00} & \pmstd{0.00}{0.00} & \pmstd{0.00}{0.00} \\
& RECT            & \pmstd{49.91}{0.15} & \pmstd{49.91}{0.38} & \pmstd{50.09}{0.28} & \pmstd{336.42}{0.24} & \pmstd{4.07}{0.09} & \pmstd{0.00}{0.00} & \pmstd{0.00}{0.00} & \pmstd{0.00}{0.00} \\
& AlphaEdit       & \pmstd{66.27}{0.34} & \pmstd{58.75}{0.20} & \pmstd{49.98}{0.17} & \pmstd{555.21}{1.06} & \pmstd{4.12}{0.12} & \pmstd{22.30}{0.32} & \pmstd{20.34}{0.15} & \pmstd{1.57}{0.03} \\
& NAS  & \best{\pmstd{97.95}{0.32}} & \best{\pmstd{82.71}{0.19}} & \best{\pmstd{60.44}{0.23}} & \best{\pmstd{619.72}{0.41}} & \best{\pmstd{31.88}{0.32}} & \best{\pmstd{93.17}{0.35}} & \best{\pmstd{88.30}{0.33}} & \second{\pmstd{32.15}{0.13}} \\

\modelsep

\multirow{10}{*}{\vmodel{Qwen2.5}}%
& Pre-edited   & \pmstd{12.58}{0.17} & \pmstd{15.26}{0.25} & \pmstd{86.24}{0.28} & \pmstd{338.63}{0.28} & \pmstd{6.87}{0.09} & \pmstd{34.83}{0.14} & \pmstd{33.76}{0.16} & \pmstd{38.60}{0.20} \\
\presep
& FT           & \second{\pmstd{88.45}{0.14}} & \best{\pmstd{69.30}{0.26}} & \pmstd{42.18}{0.13} & \pmstd{532.51}{0.20} & \pmstd{2.70}{0.03} & \pmstd{21.15}{0.35} & \pmstd{10.23}{0.19} & \pmstd{3.42}{0.10} \\
& UltraEdit      & \pmstd{63.03}{0.21} & \pmstd{55.78}{0.23} & \pmstd{40.34}{0.33} & \pmstd{575.56}{0.27} & \pmstd{18.33}{0.17} & \pmstd{65.76}{0.19} & \pmstd{62.81}{0.23} & \pmstd{28.19}{0.17} \\
& RLEdit           & \pmstd{75.50}{0.27} & \pmstd{53.83}{0.36} & \pmstd{46.09}{0.16} & \pmstd{605.33}{0.25} & \pmstd{14.29}{0.35} & \pmstd{93.75}{0.33} & \pmstd{85.37}{0.20} & \pmstd{35.61}{0.20} \\
\blocksep
& MEMIT        & \pmstd{51.27}{0.25} & \pmstd{51.28}{0.10} & \pmstd{48.73}{0.14} & \pmstd{524.64}{0.11} & \pmstd{0.78}{0.02} & \pmstd{0.00}{0.00} & \pmstd{0.00}{0.00} & \pmstd{0.00}{0.00} \\
& PRUNE        & \pmstd{53.16}{0.17} & \pmstd{53.17}{0.14} & \pmstd{46.83}{0.15} & \pmstd{531.26}{0.22} & \pmstd{0.16}{0.05} & \pmstd{0.00}{0.00} & \pmstd{0.00}{0.00} & \pmstd{0.00}{0.00} \\
& RECT         & \pmstd{51.92}{0.16} & \pmstd{52.03}{0.29} & \pmstd{48.03}{0.13} & \pmstd{512.85}{1.16} & \pmstd{0.37}{0.03} & \pmstd{0.00}{0.00} & \pmstd{0.01}{0.00} & \pmstd{0.00}{0.00} \\
& AlphaEdit    & \pmstd{84.59}{0.10} & \second{\pmstd{58.41}{0.23}} & \second{\pmstd{68.96}{0.38}} & \best{\pmstd{626.15}{0.60}} & \best{\pmstd{33.45}{0.38}} & \best{\pmstd{97.25}{0.21}} & \second{\pmstd{86.17}{0.29}} & \second{\pmstd{40.16}{0.10}} \\
& NAS          & \best{\pmstd{90.44}{0.29}} & \pmstd{55.05}{0.27} & \best{\pmstd{73.21}{0.22}} & \second{\pmstd{624.10}{0.28}} & \second{\pmstd{30.63}{0.37}} & \second{\pmstd{96.84}{0.23}} & \best{\pmstd{88.10}{0.30}} & \best{\pmstd{42.52}{0.34}} \\

\modelsep

\multirow{10}{*}{\vmodel{GPT-J}}%
& Pre-edited & \pmstd{15.12}{0.11} & \pmstd{17.55}{0.35} & \pmstd{83.71}{0.21} & \pmstd{622.18}{0.23} & \pmstd{29.68}{0.16} & \pmstd{27.23}{0.16} & \pmstd{26.43}{0.31} & \pmstd{27.23}{0.26} \\
\presep
& FT         & \pmstd{57.33}{0.25} & \second{\pmstd{54.90}{0.22}} & \pmstd{44.75}{0.23} & \pmstd{576.65}{0.10} & \pmstd{3.09}{0.07} & \pmstd{11.85}{0.27} & \pmstd{10.26}{0.17} & \pmstd{0.51}{0.01} \\
& UltraEdit      & \pmstd{56.29}{0.19} & \pmstd{49.42}{0.14} & \pmstd{51.21}{0.26} & \pmstd{498.91}{0.58} & \pmstd{14.98}{0.18} & \pmstd{65.76}{0.35} & \pmstd{62.81}{0.36} & \best{\pmstd{28.19}{0.29}} \\
& RLEdit           & \second{\pmstd{78.10}{0.35}} & \pmstd{54.86}{0.18} & \pmstd{48.78}{0.34} & \second{\pmstd{579.45}{0.33}} & \second{\pmstd{16.76}{0.36}} & \pmstd{70.51}{0.26} & \pmstd{66.69}{0.34} & \second{\pmstd{24.00}{0.29}} \\
\blocksep
& MEMIT      & \pmstd{49.47}{0.13} & \pmstd{49.54}{0.27} & \pmstd{50.54}{0.37} & \pmstd{509.95}{0.43} & \pmstd{4.71}{0.14} & \pmstd{2.01}{0.06} & \pmstd{0.14}{0.00} & \pmstd{0.14}{0.01} \\
& PRUNE      & \pmstd{48.86}{0.22} & \pmstd{49.01}{0.38} & \pmstd{50.94}{0.16} & \pmstd{448.54}{1.00} & \pmstd{4.74}{0.03} & \pmstd{1.06}{0.02} & \pmstd{0.03}{0.02} & \pmstd{0.03}{0.01} \\
& RECT       & \pmstd{48.79}{0.25} & \pmstd{48.74}{0.30} & \pmstd{51.26}{0.35} & \pmstd{240.35}{0.16} & \pmstd{4.75}{0.11} & \pmstd{0.00}{0.00} & \pmstd{0.00}{0.00} & \pmstd{0.00}{0.00} \\
& AlphaEdit  & \pmstd{62.65}{0.37} & \pmstd{54.57}{0.21} & \second{\pmstd{54.73}{0.20}} & \pmstd{508.65}{0.34} & \pmstd{2.90}{0.12} & \second{\pmstd{86.13}{0.23}} & \second{\pmstd{79.43}{0.23}} & \pmstd{21.84}{0.10} \\
& NAS        & \best{\pmstd{98.73}{0.12}} & \best{\pmstd{90.68}{0.19}} & \best{\pmstd{60.75}{0.22}} & \best{\pmstd{584.59}{1.19}} & \best{\pmstd{40.37}{0.29}} & \best{\pmstd{92.08}{0.33}} & \best{\pmstd{85.04}{0.10}} & \pmstd{22.24}{0.19} \\

\bottomrule
\end{tabular}%
}
\end{table*}
}

\newcommand{\TabExtraLLM}{
\begin{table}[H]
\centering
\small
\renewcommand{\arraystretch}{1.06}
\addtolength{\tabcolsep}{-1.5pt}
\caption{
Additional Long-Horizon Sequential model editing results on GPT2-XL. Reported values are mean$\pm$std for each metric. Pre-edited is separated by a rule; Best in \best{bold}, second-best in \second{underlined}.
}
\label{tab:seq_edit_gpt2xl_only}

\resizebox{0.96\textwidth}{!}{%
\begin{tabular}{@{}c l n n n n n n n n@{}}
\toprule
\multirow{2}{*}{\textbf{Model}} & \multirow{2}{*}{\textbf{Method}}
& \multicolumn{5}{c@{}}{\textbf{CounterFact}}
& \multicolumn{3}{@{}c}{\textbf{ZsRE}} \\
\cmidrule(r{10pt}){3-7}\cmidrule(l{10pt}){8-10}
& & \textbf{Eff.$\uparrow$} & \textbf{Gen.$\uparrow$} & \textbf{Spe.$\uparrow$}
  & \textbf{Flu.$\uparrow$} & \textbf{Consis.$\uparrow$}
  & \textbf{Eff.$\uparrow$} & \textbf{Gen.$\uparrow$} & \textbf{Spe.$\uparrow$} \\
\midrule

\multirow{6}{*}{\vmodel{GPT2-XL}}%
& Pre-edited      & \pmstd{21.34}{0.25} & \pmstd{24.07}{0.31} & \pmstd{78.72}{0.19} & \pmstd{527.26}{0.88} & \pmstd{2.86}{0.04} & \pmstd{22.59}{0.22} & \pmstd{21.80}{0.18} & \pmstd{24.33}{0.21} \\
\presep
& MEMIT           & \pmstd{49.56}{0.14} & \pmstd{49.56}{0.14} & \pmstd{50.42}{0.22} & \pmstd{496.38}{0.45} & \pmstd{4.34}{0.11} & \pmstd{0.00}{0.00} & \pmstd{0.00}{0.00} & \pmstd{0.00}{0.00} \\
& PRUNE           & \pmstd{47.86}{0.21} & \pmstd{48.14}{0.19} & \pmstd{52.72}{0.25} & \pmstd{469.10}{0.72} & \pmstd{4.32}{0.13} & \pmstd{0.00}{0.00} & \pmstd{0.00}{0.00} & \pmstd{0.00}{0.00} \\
& RECT            & \pmstd{47.66}{0.18} & \pmstd{47.47}{0.23} & \pmstd{52.32}{0.29} & \pmstd{235.52}{0.33} & \pmstd{0.70}{0.02} & \pmstd{0.00}{0.00} & \pmstd{0.00}{0.00} & \pmstd{0.00}{0.00} \\
& AlphaEdit       & \second{\pmstd{73.92}{0.28}} & \second{\pmstd{58.16}{0.24}} & \second{\pmstd{54.20}{0.31}} & \second{\pmstd{566.62}{1.02}} & \second{\pmstd{14.53}{0.27}} & \second{\pmstd{43.62}{0.33}} & \second{\pmstd{35.90}{0.29}} & \second{\pmstd{10.52}{0.11}} \\
& NAS             & \best{\pmstd{91.72}{0.18}} & \best{\pmstd{73.15}{0.25}} & \best{\pmstd{56.76}{0.22}} & \best{\pmstd{568.23}{0.95}} & \best{\pmstd{27.97}{0.31}} & \best{\pmstd{58.54}{0.26}} & \best{\pmstd{49.32}{0.28}} & \best{\pmstd{12.76}{0.15}} \\

\bottomrule
\end{tabular}%
}
\end{table}
}

\newcommand{\TabExtraMethod}{
\begin{table}[H]
\centering
\small
\renewcommand{\arraystretch}{1.06}
\addtolength{\tabcolsep}{-1.5pt}
\caption{
Sequential model editing results on additional baselines (10,000 sequential edits).
Rows are visually grouped by whether the method belongs to the L\&E family; \emph{best/second are computed across \textbf{all} methods within each backbone} (ignoring NULL entries).
Best in \best{bold}, second-best in \second{underlined}.
}
\label{tab:seq_edit_vertical_model}

\begin{tabular}{@{}c l n n n n n n n n@{}}
\toprule
\multirow{2}{*}{\textbf{Model}} & \multirow{2}{*}{\textbf{Method}}
& \multicolumn{5}{c@{}}{\textbf{CounterFact}}
& \multicolumn{3}{@{}c}{\textbf{ZsRE}} \\
\cmidrule(r{10pt}){3-7}\cmidrule(l{10pt}){8-10}
& & \textbf{Eff.$\uparrow$} & \textbf{Gen.$\uparrow$} & \textbf{Spe.$\uparrow$}
  & \textbf{Flu.$\uparrow$} & \textbf{Consis.$\uparrow$}
  & \textbf{Eff.$\uparrow$} & \textbf{Gen.$\uparrow$} & \textbf{Spe.$\uparrow$} \\
\midrule

\multirow{7}{*}{\vmodel{LLaMA3}}%
& GRACE           & \best{99.08} & 10.20 & \best{88.39} & \best{630.42} & 24.40 & 90.16 & 1.84 & 17.52 \\
& WISE            & 17.90 & 20.23 & \second{80.79} & 355.55 & 1.63 & 31.77 & 31.26 & 24.59 \\
& MEMOIR          & 90.44 & 64.84 & 54.71 & NULL & NULL & 90.15 & 87.67 & 31.78 \\
\blocksep
& ENCORE          & 92.03 & 84.44 & 56.98 & 577.27 & 26.81 & 90.86 & 87.31 & 31.51 \\
& LyapLock        & 37.40 & 26.96 & 70.38 & 601.12 & 27.32 & 77.09 & 72.22 & 28.90 \\
& NAS             & \second{98.85} & \second{85.50} & 64.62 & 619.99 & \second{29.93} & \second{92.76} & \second{88.71} & \best{32.28} \\
& NAS$^\dagger$      & 98.64 & \best{88.21} & \second{66.13} & \second{625.07} & \best{31.35} & \best{93.74} & \best{89.39} & \second{32.18} \\

\modelsep

\multirow{4}{*}{\vmodel{Qwen2.5}}%
& GRACE           & 96.05 & 16.49 & \best{84.48} & 334.05 & 6.44 & \best{98.64} & 0.85 & 20.04 \\
& WISE            & 11.76 & 36.45 & \second{83.94} & 397.35 & 3.80 & 30.91 & 32.43 & \second{25.87} \\
\blocksep
& NAS             & \second{98.22} & \second{65.48} & 76.42 & \second{621.36} & \second{29.83} & 72.15 & \second{60.09} & 18.78 \\
& NAS$^\dagger$      & \best{99.21} & \best{77.21} & 74.86 & \best{621.37} & \best{31.76} & \second{98.48} & \best{91.00} & \best{44.68} \\

\modelsep

\multirow{6}{*}{\vmodel{GPT-J}}%
& GRACE           & 99.27 & 16.40 & \second{80.81} & \best{621.62} & 29.78 & \second{93.77} & 1.62 & 17.99 \\
& MEMOIR          & 87.56 & 49.93 & 52.45 & NULL & NULL & 84.19 & 78.56 & \second{27.23} \\
& WISE            & 45.85 & 43.02 & 58.33 & 510.89 & 8.80 & 32.73 & 29.08 & 26.00 \\
\blocksep
& LyapLock        & 47.56 & 34.60 & \best{83.05} & 591.39 & 30.82 & 64.23 & 58.62 & \best{29.25} \\
& NAS             & \second{99.60} & \second{93.36} & 67.00 & \second{611.39} & \second{39.57} & 92.74 & \second{81.28} & 24.32 \\
& NAS$^\dagger$      & \best{99.61} & \best{93.92} & 67.19 & 609.90 & \best{42.25} & \best{97.65} & \best{92.78} & 24.60 \\

\modelsep

\multirow{7}{*}{\vmodel{GPT2-XL}}%
& GRACE           & \best{99.14} & 16.92 & \best{77.18} & 526.09 & 3.03 & \second{92.73} & 1.39 & 25.15 \\
& MELO            & 61.48 & 47.15 & 53.92 & 542.73 & 11.63 & 68.15 & 69.65 & 19.56 \\
& WISE            & 33.12 & 33.48 & 67.27 & 541.08 & 6.20 & 37.52 & 39.23 & \second{40.27} \\
\blocksep
& ENCORE          & 92.92 & 77.09 & 57.79 & 514.90 & 23.40 & 86.40 & \second{77.33} & 24.66 \\
& LyapLock        & 31.80 & 28.68 & \second{76.81} & \best{621.09} & 31.29 & 34.76 & 32.73 & 25.82 \\
& NAS             & 96.98 & \second{82.03} & 59.24 & 557.40 & \second{33.18} & 72.38 & 64.64 & 16.93 \\
& NAS$^\dagger$      & \second{97.24} & \best{83.77} & 59.54 & \second{566.36} & \best{35.32} & \best{97.07} & \best{87.74} & \best{43.18} \\

\bottomrule
\end{tabular}
\end{table}
}

\theoremstyle{plain}
\newtheorem{theorem}{Theorem}[section]
\newtheorem{proposition}[theorem]{Proposition}
\newtheorem{lemma}[theorem]{lemma}
\newtheorem{corollary}[theorem]{Corollary}
\theoremstyle{definition}
\newtheorem{definition}[theorem]{Definition}

\theoremstyle{remark}

\usepackage[disable,textsize=tiny]{todonotes}

\title{Norm Anchors Make Model Edits Last}

\author{%
Mingda Liu\thanks{Equal contribution.} \\
Institute of Science Tokyo \\
\texttt{puppetsasya@gmail.com}
\And
Zhenghan Zhu\footnotemark[1] \\
Institute of Science Tokyo
\And
Ze'an Miao\footnotemark[1] \\
Institute of Science Tokyo
\And
Katsuki Fujisawa \\
Institute of Science Tokyo \\
\texttt{fujisawa.k.2110@m.isct.ac.jp}
}

\newcommand{\StdToMeanGap}{\mkern0mu} 
\newcommand{\PMToStdGap}{\kern0.01em}          
\newcommand{\counterfactStreamSize}{20{,}877}
\newcommand{\zsreStreamSize}{19{,}086}

\newcommand{\pmstd}[2]{%
  \ensuremath{%
    #1_{\StdToMeanGap\smash{\raisebox{0.22ex}{\hbox{$\scriptscriptstyle \pm\PMToStdGap #2$}}}}%
  }%
}

\newcommand{\best}[1]{%
  \begingroup
  \setbox0=\hbox{#1}%
  \hbox{%
    \rlap{\pdfrender{TextRenderingMode=2,LineWidth=0.6pt}{\copy0}}%
    \copy0%
  }%
  \endgroup
}

\newcommand{\SecondULDepth}{0.5ex}   
\newcommand{\SecondULThick}{0.35pt}   

\newcommand{\second}[1]{%
  \begingroup\leavevmode
  \setbox0=\hbox{#1}
  \rlap{\raisebox{-\SecondULDepth}{\rule{\wd0}{\SecondULThick}}}
  \box0
  \endgroup
}

\newcolumntype{n}{>{\centering\arraybackslash}c}

\newcommand{\presep}{%
  \addlinespace[2pt]%
  \cmidrule(lr){2-10}
  \addlinespace[2pt]%
}
\newcommand{\blocksep}{%
  \addlinespace[2pt]%
  \cmidrule(lr){2-10}%
  \addlinespace[2pt]%
}
\newcommand{\modelsep}{\midrule\midrule}

\newcommand{\vmodel}[1]{\rotatebox{90}{\strut\small #1}}
\newcommand{\icmlsinglecolumnwidth}{0.48\textwidth}
\newenvironment{icmlsinglecolumnfigure}{%
  \begin{minipage}[t]{\icmlsinglecolumnwidth}\vspace{0pt}\centering
}{%
  \end{minipage}%
}

\begin{document}

\maketitle

\begin{abstract}
Sequential Locate-and-Edit (L\&E) model editing can fail abruptly after many edits. We identify and formalize this failure as a positive \emph{norm-feedback loop}, in which solved value vectors and edited MLP weights progressively amplify each other, degrading edit quality and eventually collapsing model capabilities. Our analysis shows that this feedback can yield approximately exponential norm growth under standard L\&E dynamics, and can remain unresolved by existing increment-level regularizers or update clamps. We propose \textbf{Norm-Anchor Scaling (NAS)}, a plug-in stabilizer that breaks this loop by rescaling each solved value vector to an original-model reference norm. Across multiple LLM backbones, datasets, and L\&E editors, NAS extends the usable editing horizon by more than \textbf{4$\times$} and improves long-run editing performance by \textbf{72.2\%} on average, while preserving single-edit efficacy, with only a \textbf{one-line modification and negligible computational overhead}. The code is available at \url{https://github.com/SasyaTitech/NAS}.
\end{abstract}

\section{Introduction}
\label{sec:intro}

Large language models (LLMs) store substantial factual knowledge~\citep{petroni-etal-2019-language,roberts-etal-2020-much}, 
but deployed models inevitably encounter outdated, incorrect, or newly emerging facts~\citep{lin-etal-2022-truthfulqa,de-cao-etal-2021-editing,dhingra-etal-2022-time}. Correcting such errors by full retraining is prohibitively expensive~\citep{de-cao-etal-2021-editing}, while broad fine-tuning can modify unrelated behavior in hard-to-predict ways~\citep{zhang2024dissecting}. \textbf{Knowledge editing} addresses this problem by locally modifying a model so that a targeted prompt produces a desired new fact while unrelated behaviors are preserved~\citep{de-cao-etal-2021-editing,mitchell2022mend}.

Among editing approaches, \textbf{Locate-and-Edit (L\&E)}~\citep{meng2022rome, meng2023memit} methods have become a widely used in-weight editing paradigm. Given a factual update, they identify an FFN site and apply a localized low-rank update to the output projection so that a target key maps to a new value. This structure makes L\&E computationally efficient and often more localized than large-subset fine-tuning~\citep{Wangsurvey,yaosurvey}.

A practical editor, however, must be reusable across a stream of updates~\citep{Gupta2024ModeleA}. 
In this standard sequential editing setting, stability is typically assessed by whether the editor can maintain edit quality, preserve unrelated and general-purpose behavior, and avoid abrupt collapse as edits accumulate~\citep{meng2023memit,zhang2025queueeditstructuralselfcorrectionsequential}. 
Under long-stream evaluations, many existing L\&E editors exhaust their usable editing lifespan within only a few thousand sequential edits, making it a central bottleneck to extend the effective horizon of in-weight editing.

\begin{figure}[!htbp]
  \centering
  \vspace{-1mm}
  \includegraphics[width=0.93\textwidth]{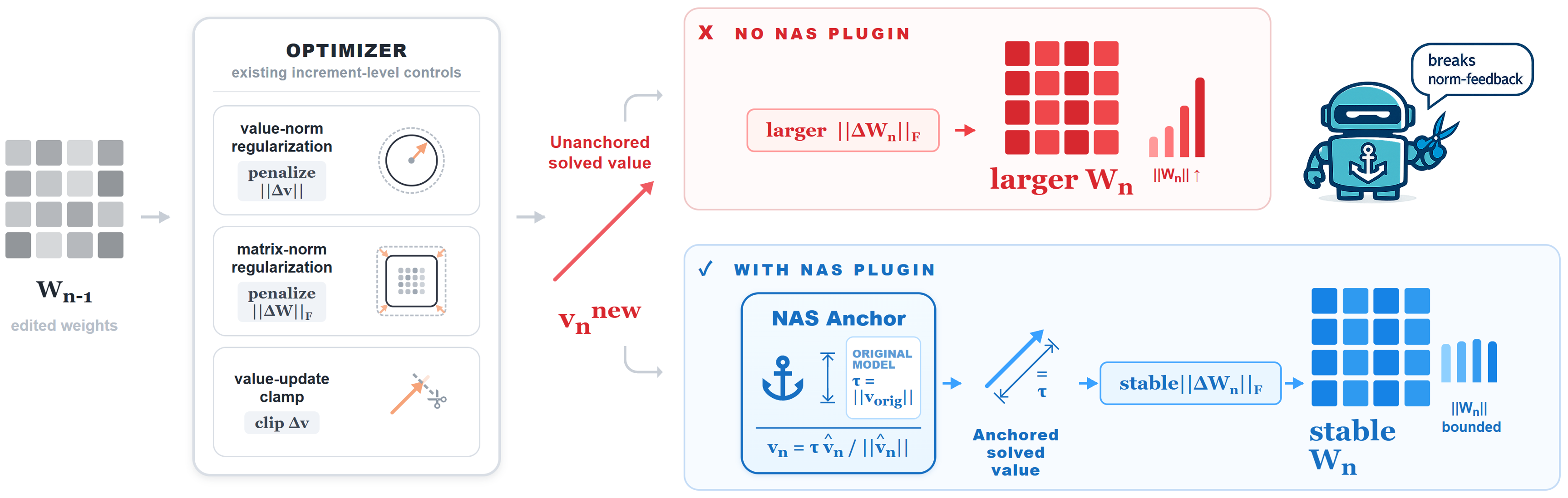}
  \vspace{-3mm}
  \caption{\textbf{Mechanism illustration.}
  NAS anchors the solved value before writing, breaking the norm-feedback loop left unresolved by increment-level controls.}
  \label{fig:nas_mechanism}
  \vspace{-2mm}
\end{figure}

\begin{wrapfigure}{r}{\icmlsinglecolumnwidth}
  \centering
  \vspace{-0.5mm}
  \includegraphics[width=\linewidth]{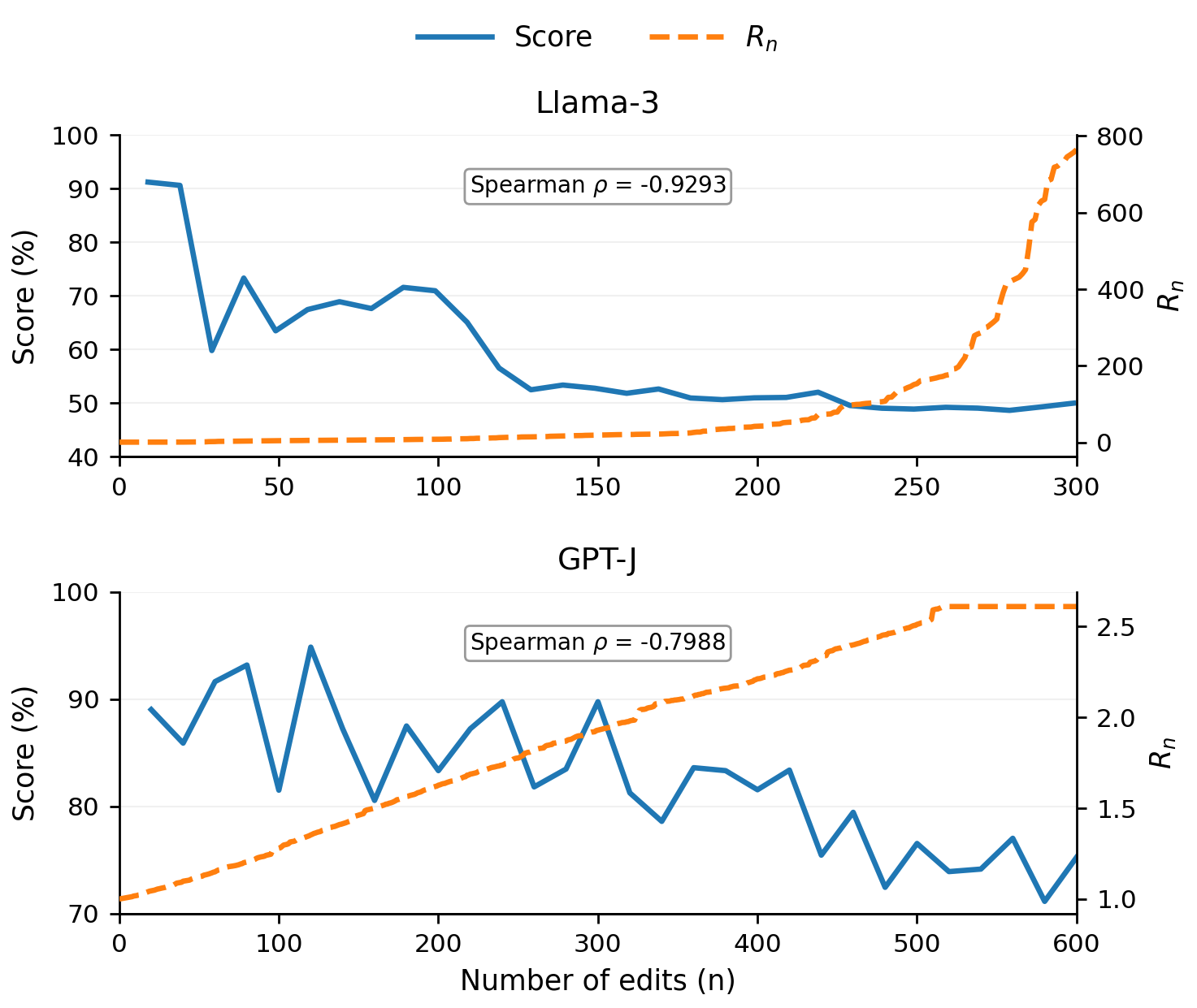}
  \vspace{-5.0mm}
  \caption{\textbf{Norm explosion accompanies sequential editing collapse.}
    During sequential MEMIT editing, we track edit success (blue, left axis) and normalized weight growth $R_n \coloneqq \| W_n\|/\| W_0\|$ (orange, right axis) versus edit step $n$.
    For both Llama-3 and GPT-J, increasing $R_n$ coincides with deteriorating editing performance;
    Spearman $\rho$ is shown in each panel.
    }
  \label{fig:accompany}
  \vspace{-0.7mm}
\end{wrapfigure}

Prior work has observed that such long-run degradation often accompanies abnormal growth in edited-parameter norms, and Fig.~\ref{fig:accompany} shows the same signature in our setting. 
We go beyond this phenomenon by formalizing the L\&E update dynamics and identifying a self-reinforcing \emph{norm-feedback instability}: past edits enlarge the edited matrix, causing later edits to solve value vectors at a larger scale, whose rank-one writes further amplify the same matrix. 
This analysis derives conditions under which the edited-matrix norm grows approximately exponentially, matching the log-linear norm trajectories observed in our experiments. 
The analysis also explains why existing increment-level norm controls, such as update regularization or clipping, can still leave the feedback unresolved: they constrain the update step but do not anchor the final solved value repeatedly written into the model (Fig.~\ref{fig:nas_mechanism}).

Motivated by this diagnosis, we propose \textbf{Norm-Anchor Scaling (NAS)}, a plug-and-play minimal stabilizer for the L\&E paradigm. 
Before applying each edit, NAS rescales the final solved value vector to an original-model reference norm while preserving its direction. 
By preventing abnormal magnitude inflation, NAS keeps the write signal within a numerically stable operating range for long-horizon sequential updates.
Our empirical-scaling analysis derives that under NAS, the target-layer weight norm admits a finite theoretical upper bound over long edit sequences.
Empirically, compared to non-NAS updates, NAS confines the target-layer norm drift to a much tighter band around its pre-edit value (Fig.~\ref{fig:rq3_ts_and_bars}), and substantially suppresses the drift of hidden representations (Fig.~\ref{fig:pca_drift_insets}).

NAS is a \textbf{one-line drop-in with negligible computational overhead} (Appendix~\ref{app:runtime}). 
Extensive experiments against a broad set of baselines, including stability-oriented L\&E editors with their own norm-related or statistical controls, show that NAS significantly delays the degradation point (by more than $\boldsymbol{4}\boldsymbol{\times}$ on average) and improves sequential editing performance: the average editing success increases from \textbf{51.9 to 89.3 (\,+37.4 percentage points; +72.2\% relative\,)}, while preserving single-edit efficacy under stress test.
Under our long-horizon \emph{atomic sequential editing} (one request per step) on the full ZsRE (\textbf{\zsreStreamSize} edits) and CounterFact (\textbf{\counterfactStreamSize} edits) streams, NAS is the \emph{only} evaluated editing strategy that does not exhibit a clear degradation.

\vspace{-1mm}
\section{Preliminaries}
\label{sec:prelim}

\subsection{Knowledge Editing Setup}
\label{sec:prelim:setup}

Factual knowledge in LLMs is often described as a triple $(s,r,o)$ (subject, relation, object), which can be queried by a prompt
$x=\mathrm{Prompt}(s,r)$ such that the model should generate $o$.
A knowledge editing request specifies a target object $o^\star$ for the same prompt $x$, and aims to modify a localized set of parameters
so that the edited model assigns high probability to $o^\star$ under $x$ while preserving unrelated behaviors.

\subsection{Transformer FFN as Key--Value Memory}
\label{sec:prelim:alm}

We focus on the FFN module at a fixed layer $l$.
Following the key--value interpretation of FFNs~\citep{geva2021kvm}, we view the intermediate activation as a
\emph{key} $k$ that matches factual patterns, while the FFN output is a retrieved \emph{value} $v$ that encodes
the information to be written into the hidden state.
Let $W_{\mathrm{out}}^{(l)}$ denote the FFN output projection; then
\begin{equation}
v \;=\; W_{\mathrm{out}}^{(l)}\,k.
\label{eq:kv_basic}
\end{equation}
For simplicity, we write $W \coloneqq W_{\mathrm{out}}^{(l)}$ henceforth.

\subsection{Locate-and-Edit Paradigm}
\label{sec:prelim:le}

Locate-and-Edit (L\&E) methods update a localized FFN matrix $W$ to rewrite a specific factual association.
After selecting an editing site, we obtain a key vector $k_n^\star$ that represents the query (e.g., the pair $(s,r)$)
at that site. The pre-edit value is
\begin{equation}
v^{\mathrm{old}} \;\coloneqq\; Wk_n^\star.
\label{eq:v_old}
\end{equation}

\paragraph{Overall update objective.}
The core objective is to modify $W$ so that the target key $k_n^\star$ maps to a desired value $v^{\mathrm{new}}$,
while keeping the original input--output behavior of $W$ on typical keys as unchanged as possible:
\begin{equation}
\min_{W'}\; \mathbb{E}_{k\sim\mathcal{D}}\,\|W'k - Wk\|^2
\quad \text{s.t.}\quad W'k_n^\star = v^{\mathrm{new}},
\label{eq:le_objective}
\end{equation}
where $\mathcal{D}$ denotes a text-induced distribution over FFN keys.

\paragraph{Closed-form rank-one update of $W$.}
Let $C \coloneqq \mathbb{E}_{k\sim\mathcal{D}}[kk^\top]$ be the (pre-computed) second-moment matrix of keys.
Following prior L\&E methods, the constrained problem~\eqref{eq:le_objective} admits a closed-form rank-one solution
\citep{meng2022rome}. Define the per-edit update $\Delta W$ as
\begin{equation}
\Delta W
\;\coloneqq\;
\big(v^{\mathrm{new}} - v^{\mathrm{old}}\big)\,
\frac{(C^{-1}k_n^\star)^\top}{(k_n^\star)^\top C^{-1}k_n^\star}.
\label{eq:deltaW_closed_form}
\end{equation}
Then the updated matrix is
\begin{equation}
W' \;=\; W + \Delta W,
\label{eq:W_update_deltaW}
\end{equation}
which enforces $W'k_n^\star = v^{\mathrm{new}}$ while minimizing the expected disturbance
$\mathbb{E}_{k\sim\mathcal{D}}\,\|W'k - Wk\|^2$.

\paragraph{Computing the target value $v^{\mathrm{new}}$.}
The target value for the new fact is obtained by optimizing an additive
correction $\Delta \in \mathbb{R}^{d}$ under the edit prompt
$x=\mathrm{Prompt}(s,r)$:
\begin{equation}
    \Delta \in \arg\min_{\Delta \in \mathbb{R}^{d}}
    \mathcal{L}_{\mathrm{NLL}}\!\bigl(o^\star \mid x;\, v^{\mathrm{old}}+\Delta\bigr),
    \label{eq:delta_opt}
\end{equation}
where $\mathcal{L}_{\mathrm{NLL}}$ is evaluated with the FFN output at the editing site
replaced by $v^{\mathrm{old}}+\Delta$. The target value is then defined as
\begin{equation}
    v^{\mathrm{new}} \coloneqq v^{\mathrm{old}} + \Delta .
    \label{eq:vnew_delta}
\end{equation}

\subsection{Sequential Editing}
\label{sec:prelim:seq}

Sequential editing applies a stream of requests $\{(s_n,r_n,o_n^\star)\}_{n=1}^{T}$ to the same base model.
Focusing on the edited matrix $W$, the parameter trajectory follows
\begin{equation}
W_{0} = W_{\mathrm{base}}, \qquad
W_{n} = W_{n-1} + \Delta W_{n}, \qquad n=1,\dots,T.
\label{eq:W_n_and_W_n-1}
\end{equation}

For the $n$-th request, the pre-edit value is given by
\begin{equation}
v_{n}^{\mathrm{old}} \;=\; W_{n-1}k_n^\star.
\label{eq:vold_n}
\end{equation}

and the closed-form per-edit update is
\begin{equation}
\Delta W_{n}
\;\coloneqq\;
\big(v_{n}^{\mathrm{new}} - v_{n}^{\mathrm{old}}\big)\,
\frac{(C^{-1}k_n^\star)^\top}{(k_n^\star)^\top C^{-1}k_n^\star}.
\label{eq:deltaW_n}
\end{equation}
In the next section, we analyze the induced dynamics of $\|W_{n}\|^2$ and relate them to the norm statistics
of $v_{n}^{\mathrm{old}}$ and $v_{n}^{\mathrm{new}}$.

\section{Analysis and Method}

\subsection[Computing Wn norm squared]{Computing $\|W_n\|^2$}
\label{sec:analysis:wnorm}

In this section, we conduct both the analysis and all sequential-editing experiments under the Euclidean metric.
We focus on the most basic locate-and-edit (L\&E) update without additional statistical constraints, and thus set $C=I$ for clarity.
The general $C\neq I$ case follows analogously; see Appendix~\ref{app:proofgeneral}.
Under this setting, Eq.~\eqref{eq:deltaW_n} becomes:
\begin{equation}
\label{eq:deltaW_n_CI}
\Delta W_n
\;=
\frac{\bigl(v_n^{\mathrm{new}}-v_n^{\mathrm{old}}\bigr){k_n^\star}^\top}{\| k_n^\star\|^2}.  
\end{equation}

Combining Eqs.~\eqref{eq:W_n_and_W_n-1} and~\eqref{eq:deltaW_n_CI} yields an explicit recursion for $\| W_n\|^2$.

\begin{lemma}
\label{lemma3.1}
\label{lem:norm_recursion}
The squared norm of the edited weight matrix satisfies
\begin{equation}
\label{eq:lemma3.1}
\|W_n\|^2
=
\|W_{n-1}\|^2
+
\frac{\| v_n^{\mathrm{new}}\|^2-\| v_n^{\mathrm{old}}\|^2}{\| k_n^\star\|^2}.
\end{equation}
\end{lemma}
\noindent
The proof is deferred to Appendix~\ref{app:proof3.1}.

\paragraph{Empirical scaling assumptions.}
We use the following empirical relations: $\|k_n^\star\|^{-2}\approx K$,
\[
\mathbb{E}\!\left[\|v_n^{\mathrm{new}}\|^2 \mid W_{n-1}\right]
\approx s_{\mathrm{new}}\|W_{n-1}\|^2+b_{\mathrm{new}},\quad
\mathbb{E}\!\left[\|v_n^{\mathrm{old}}\|^2 \mid W_{n-1}\right]
\approx s_{\mathrm{old}}\|W_{n-1}\|^2+b_{\mathrm{old}}.
\]
with $s_{\mathrm{new}}>s_{\mathrm{old}}>0$. Appendix~\ref{app:proof3.2} empirically validates these relations and reports the supporting fits (Figs.~\ref{fig:stable_k} and~\ref{fig:Ev-W}).

\paragraph{Exponential growth of $\| W_n\|^2$.}
Under Lemma~\ref{lem:norm_recursion} and the empirical scaling assumptions above,
taking expectation on both sides of Eq.~\eqref{eq:lemma3.1}
yields the following consequence.

\begin{proposition}
\label{prop3.2}
Under the empirical scaling assumptions above, there exist constants $\alpha\in \mathbb{R}$ and $R>1$ such that
\begin{equation}
\label{eq:prop3.2}
\mathbb{E}\| W_n\|^2
\approx
R^n\,\mathbb{E}\| W_0\|^2\
+
\alpha\,(R^n-1).
\end{equation}
Consequently, $\mathbb{E}\| W_n\|^2$ grows exponentially with the number of edits.
\end{proposition}

\begin{wrapfigure}{r}{\icmlsinglecolumnwidth}
  \centering
  \vspace{-0.4\baselineskip}
  \includegraphics[width=0.94\linewidth]{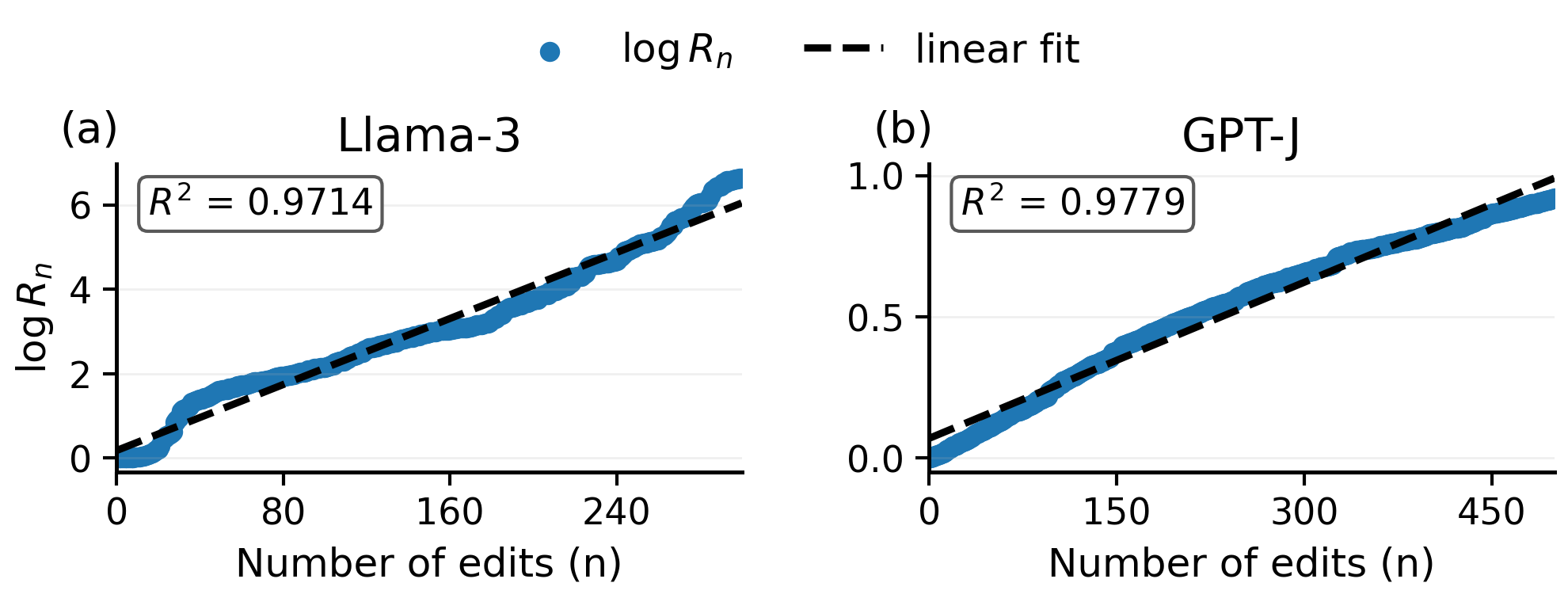}
  \vspace{-0.5mm}
  \caption{\textbf{Log-linear growth of weight-norm ratio under sequential editing.}
  We measure $\log R_n$ as a function of edit step $n$ ($R_n=\| W_n\|/\| W_0\|$) for Llama-3 and GPT-J using MEMIT. Linear fits (dashed) achieve high $R^2$, supporting an approximately exponential increase of $R_n$ with the number of edits.}
  \label{fig:logRn_linearfit}
  \vspace{-2.2\baselineskip}
\end{wrapfigure}

\noindent
The proof is deferred to Appendix ~\ref{app:proof3.2}.

\paragraph{Verification via curve fitting.}
We fit $\| W_n\|$ as a function of $n$ and observe a clear exponential trend (Fig.~\ref{fig:logRn_linearfit}).
In particular, a linear fit on $\log(\cdot)$-scaled trajectories corroborates the predicted exponential growth.

\subsection{Method}
\label{sec:method}

\paragraph{Motivation.}
Section~\ref{sec:analysis:wnorm} suggests that under unconstrained locate-and-edit (L\&E) updates,
the edited weight norm can grow rapidly with the number of edits.
Empirically, $\| v_n^{\mathrm{new}}\|^2$ increases with the current state
(e.g., $\|W_{n-1}\|^2 $; Fig.~\ref{fig:Ev-W}),
forming a positive feedback loop: larger $\| W\|^2$ induces larger target values, which in turn further
amplify $\| W\|^2$ through the rank-one updates.

\paragraph{Norm anchoring via rescaling.}
We break this feedback by explicitly controlling the magnitude of the injected value vector.
Let
\begin{equation}
\label{eq:anchor_tau_def}
\tau \;\coloneqq\; \bigl(\mathbb{E}\| v^{\mathrm{new}}\|^2\bigr)^{1/2} > 0\;\;\text{measured on an original (unedited) model},
\end{equation}
where the expectation is estimated by performing pilot edits on $N$ randomly sampled facts and averaging the
resulting $\| v^{\mathrm{new}}\|^2$ (Implementation details and ablation of N in Appendix~\ref{app:implementNAS}).

For each edit step $n$, suppose the base editor produces an unconstrained
target value $\hat v_n^{\mathrm{new}}$. We then rescale it as
\begin{equation}
\label{eq:anchor_rescale}
v_n^{\mathrm{new}}
\;\leftarrow\;
\frac{\tau}{\| \hat v_n^{\mathrm{new}}\|}
\hat v_n^{\mathrm{new}},
\end{equation}
which enforces $\| v_n^{\mathrm{new}}\| = \tau$ by construction.
Intuitively, this introduces negative feedback: as the system drifts and the unconstrained
$\| \hat v_n^{\mathrm{new}}\|^2$ grows, the rescaling factor shrinks accordingly.

\paragraph{Implication for weight-norm dynamics.}
We next show that norm anchoring removes the divergence predicted by the unconstrained analysis.

\begin{corollary}
\label{cor3.3}
If we implement Eq.~\eqref{eq:anchor_rescale} for each $n\in \{1,2,...,T\}$, then there exist constants $\beta >0$, $r \in(0,1)$ such that
\begin{equation}
\label{eq:cor3.3}
\mathbb{E}\|W_n\|^2
\approx
r^n\,\mathbb{E}\| W_0\|^2
+
\beta\,(1-r^n).
\end{equation}
Consequently, $\mathbb{E}\|W_n\|^2$ remains bounded as $n\to\infty$.
\end{corollary}
\noindent
The proof is deferred to Appendix~\ref{app:proof3.3}.

\begin{center}
\begin{minipage}[t]{\icmlsinglecolumnwidth}
\vspace{0pt}
\paragraph{Hidden representation drift.}
To probe activation stability under long sequential updates, we sample target-layer representations on 1{,}000
\emph{held-out} factual prompts that are disjoint from the edit stream, from three model states:
\textit{pre-edited}, \textit{vanilla} L\&E, and \textit{vanilla}+NAS.
Here, \textit{vanilla} denotes MEMIT without NAS.
At 100 and 500 cumulative edits, we project representations into a shared 2D PCA space.
Figure~\ref{fig:pca_drift_insets} shows a clear failure mode of vanilla L\&E: as edits accumulate, representations
exhibit a substantial centroid shift and markedly increased dispersion relative to the pre-edited distribution,
and the drift amplifies sharply from 100 to 500 edits.
In contrast, NAS-stabilized updates keep the representation cloud tightly aligned with the pre-edited distribution.
This observation suggests that sequential degradation is accompanied by an out-of-distribution (OOD) shift in target-layer activations,
which can cause downstream layers to operate outside their pre-training regime and thereby trigger broad behavioral failures.
We next quantify these effects under long-horizon sequential editing in Section \ref{sec:experiments}.
		\end{minipage}\hfill
	\begin{icmlsinglecolumnfigure}
	  \vspace{-4\baselineskip}
	  \includegraphics[width=\linewidth]{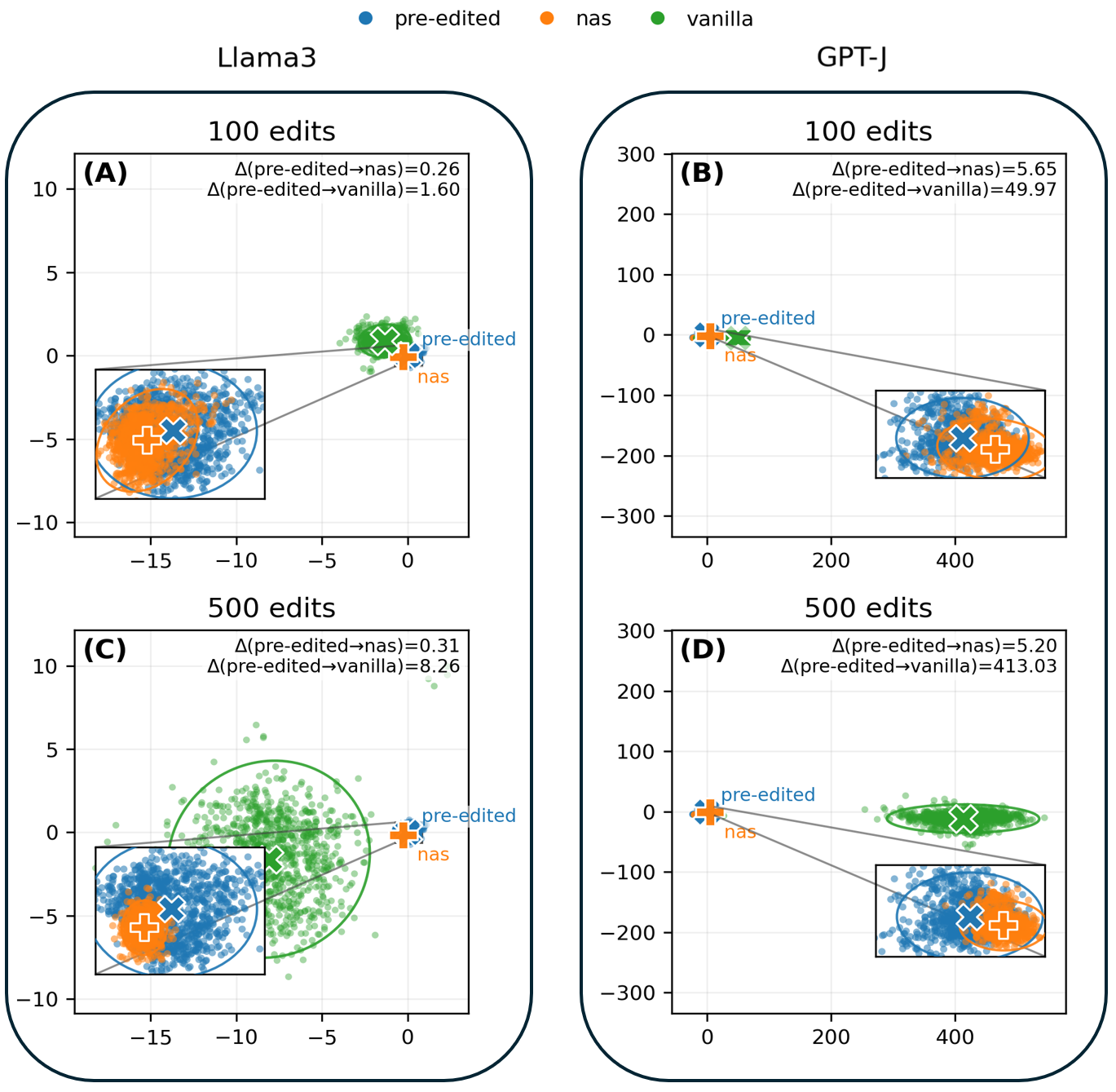}
		\captionof{figure}{
Hidden representation drift under sequential L\&E updates.
We probe the target-module representations on 1{,}000 held-out factual prompts for \textit{pre-edited} (blue),
\textit{vanilla} (green), and \textit{vanilla}+NAS (orange) after 100 and 500 edits, and visualize them in a shared 2D PCA space (PCA fit on \textit{pre-edited}).
Cross markers indicate state-wise means; ellipses denote 95\% confidence regions.
We report $\Delta(\textit{pre}\!\to\!\cdot)=\|\mu_{\cdot}-\mu_{\textit{pre}}\|_2$ (centroid distance in the original hidden space).
}
  \label{fig:pca_drift_insets}
  \vspace{-0.2mm}
\end{icmlsinglecolumnfigure}
\end{center}

\vspace{-1.5\baselineskip}
\section{Experiments}
\label{sec:experiments}

In this section, we evaluate NAS and answer the following research questions (RQs):
\begin{itemize}
  \item \textbf{RQ1.} Across multiple backbone models and datasets under long-horizon \emph{atomic sequential editing}, does NAS achieve superior editing performance and stronger stability compared to existing methods?
  \item \textbf{RQ2.} During long-horizon sequential editing, does NAS better preserve the model's general capability than existing methods, thereby mitigating degradation on general-purpose tasks?
  \item \textbf{RQ3.} As a plug-and-play component that requires only \emph{a single line of code} to integrate, can NAS suppress norm explosion when attached to different Locate-and-Edit (L\&E) baselines, while simultaneously improving editing performance and overall model behavior?
  \item \textbf{RQ4.} Does Norm Anchoring Trade Single-Edit Efficacy for Long-Horizon Stability?

\end{itemize}

\TabMainResults

\begin{figure*}[t]
  \vspace{-4mm}
  \centering
  \includegraphics[width=\textwidth]{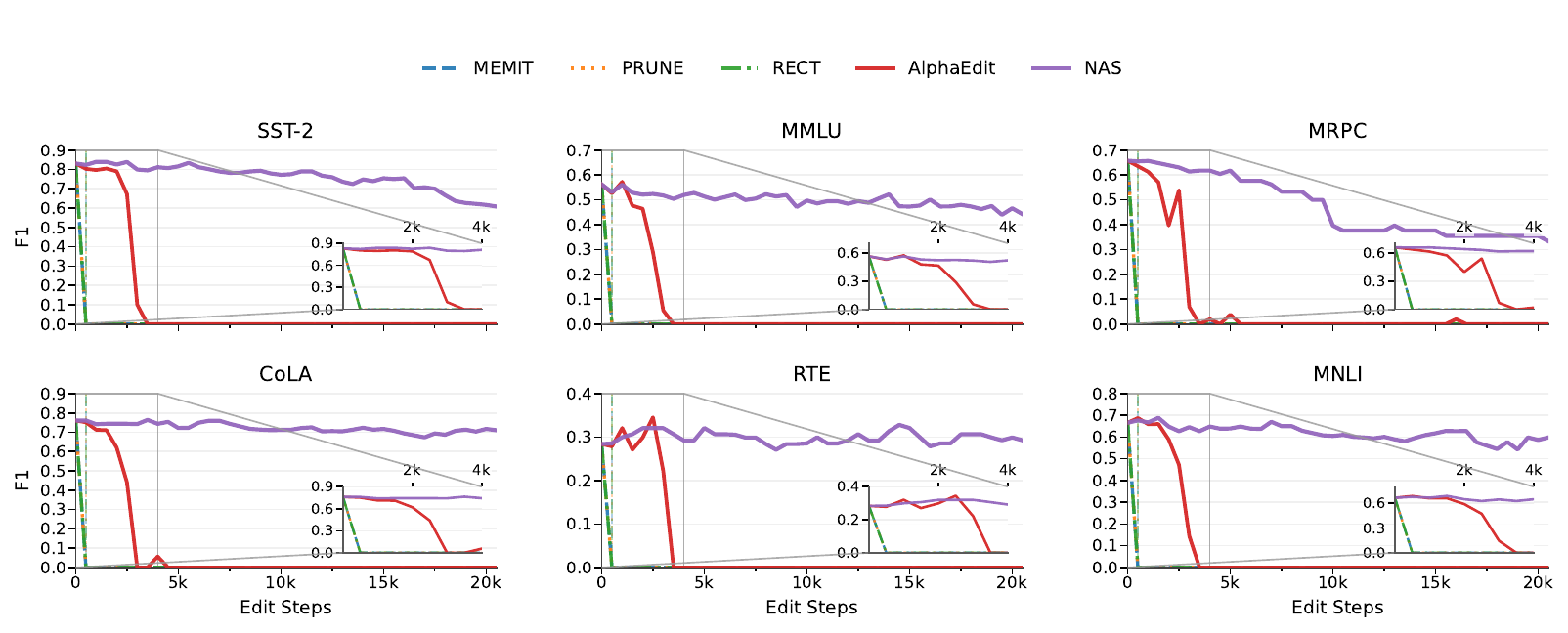}
  \caption{RQ2: GLUE performance (F1) during sequential editing. NAS preserves base-task performance substantially longer than baselines. Insets zoom into the early-edit region (0--4k); inset x-ticks are shown at 2k and 4k for readability.}
  \label{fig:rq2_glue_f1_seq}
  \vspace{-5mm}
\end{figure*}

\begin{figure*}[t]
  \centering
  \vspace{-1mm}
  \includegraphics[width=\textwidth]{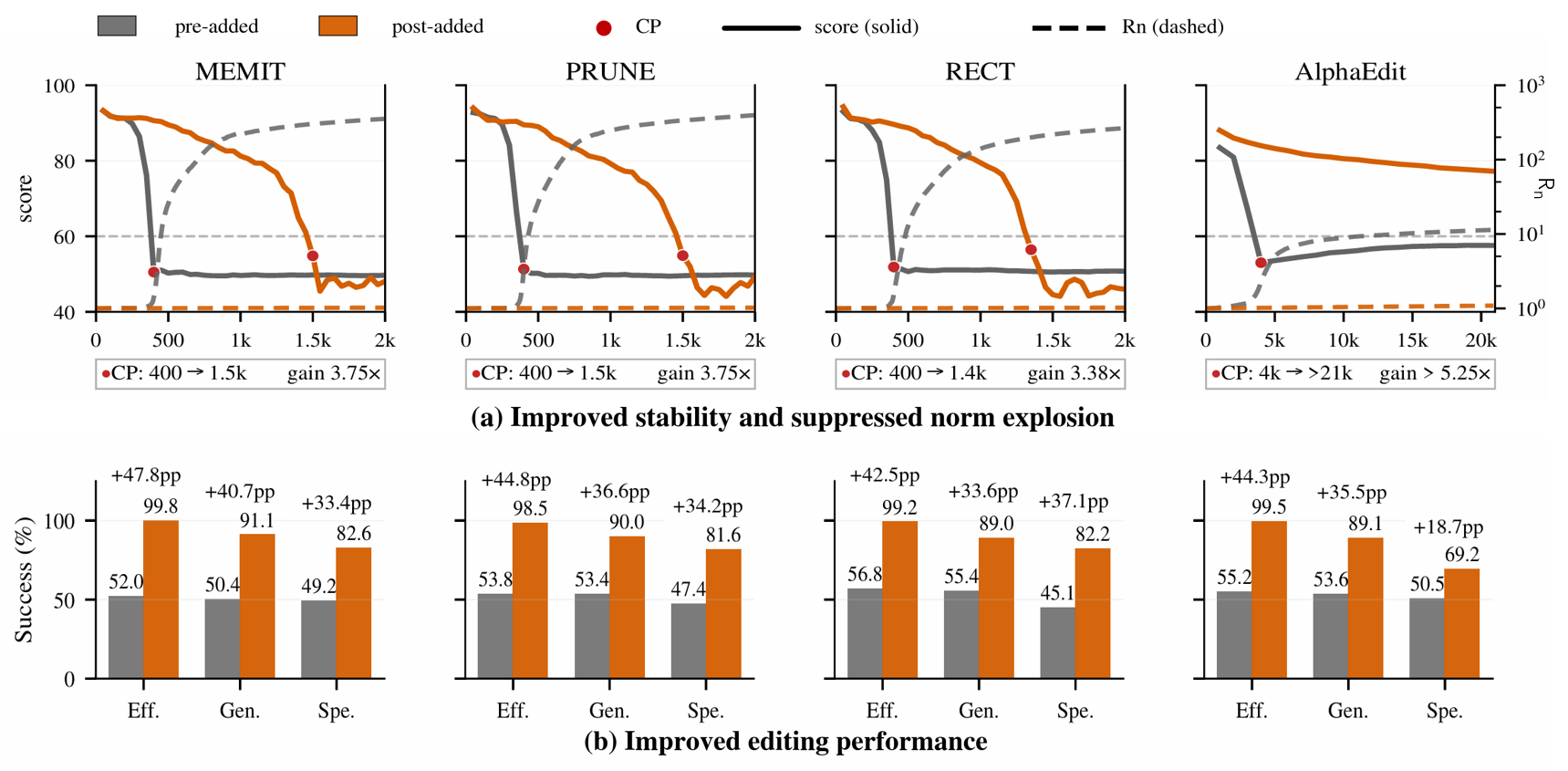}
  \caption{
  RQ3: \textbf{NAS improves stability and editing performance under long-horizon sequential editing.}
  \textbf{(a)} Dual-axis trajectories: solid lines show post-edit score, dashed lines show $R_n$; the red dot marks the collapse point (CP).
  \textbf{(b)} Editing quality at baselines' CP, reported as Eff./Gen./Spe. Comparing pre-added (gray) vs.\ post-added (orange).
  }
  \label{fig:rq3_ts_and_bars}
  \vspace{-3mm}
\end{figure*}

\subsection{Experimental Setup}
\label{sec:exp_setup}

This section briefly describes the datasets, evaluation metrics, backbone models, and baseline methods used in our experiments.
Additional details and extended settings are deferred to Appendix~\ref{apA}.

\paragraph{Backbone Models.}
Our main experiments use Llama3-8B~\citep{llama3}, Qwen2.5-7B~\citep{qwen2025qwen25technicalreport}, and GPT-J~\citep{gpt-j}.
As additional backbones, we also conduct supplementary evaluations on GPT2-XL~\citep{gpt2-xl} (Appendix~\ref{app:extrallm}).

\paragraph{Datasets and Metrics.} We adopt two widely used knowledge editing benchmarks, CounterFact~\citep{meng2022rome} and ZsRE~\citep{zsrelevy}, and follow standard protocols to report \textit{Efficacy}, \textit{Generalization}, \textit{Specificity}, \textit{Fluency}, \textit{Consistency}, and \textit{Score}. To characterize long-horizon failure under sequential editing, we additionally report the \textit{collapse point} (CP@60) (see Appendix~\ref{app:metrics:counterfact}). Beyond the main CounterFact and ZsRE evaluations, we report auxiliary benchmarks in the appendix. 
WikiBigEdit~\citep{thede2025wikibigedit} provides a large-scale stream of real-world Wikidata edits for longer-horizon evaluation and additional L\&E ablations (Appendix~\ref{app:wikibigedit}). 
MQuAKE-CF-3k-v2~\citep{zhong-etal-2023-mquake} probes counterfactual multi-hop editing (Appendix~\ref{app:multihop}), while WikiRecent and WikiBio from KnowEdit~\citep{zhang2024comprehensive} provide shorter recent and biographical factual-update streams for breadth checks (Appendix~\ref{app:short_knowedit}).

\paragraph{Baselines.}
We evaluate a diverse set of baselines spanning both \emph{non-L\&E} and \emph{L\&E} paradigms.
For non-L\&E in-weight editors, we include FT~\citep{ft2020}, 
as well as the recent lifelong editors UltraEdit~\citep{gu2025ultraedittrainingsubjectmemoryfree} and RLEdit~\citep{li2025rledit}.
For L\&E methods, we include MEMIT~\citep{meng2023memit}, PRUNE~\citep{ma2025prune}, RECT~\citep{gu2024rect},
and AlphaEdit~\citep{fang2025alphaedit}.
In Appendix~\ref{app:extrabaselines}, we report results for recent competitive L\&E methods ENCORE~\citep{gupta2025norm15k} and LyapLock~\citep{wang2025lyaplockboundedknowledgepreservation}, as official hyperparameters for our main backbones are unavailable.
We also report representative \emph{backbone-frozen} editors in Appendix~\ref{app:extrabaselines} , including GRACE~\citep{hartvigsen2023grace}, MELO~\citep{yu2024melo}, WISE~\citep{wang2024wise}, and MEMOIR~\citep{wang2025memoirlifelongmodelediting}.

\textbf{General language benchmarks.}
To assess how different editing methods affect the base capabilities of the models, we measure performance on standard language understanding benchmarks during large number of sequential edits.
We include GLUE-style tasks~\citep{glue} such as SST~\citep{sst}, MRPC~\citep{mrpc}, RTE~\citep{rte}, CoLA~\citep{cola}, and MNLI~\citep{mnli}, and MMLU~\citep{mmlu} (see Appendix~\ref{app:general_capability} for details).

\subsection{Long-horizon Sequential Editing(RQ1)}
\label{sec:exp:rq1}

We evaluate editors under \emph{atomic sequential editing} on the \emph{full} CounterFact (\counterfactStreamSize{} edits) and ZsRE (\zsreStreamSize{}
edits) streams: each request edits exactly one fact and immediately commits a weight update before proceeding to the
next. Main results are summarized in Table~\ref{tab:seq_edit_long}; results on
GPT-2XL are provided in Appendix~\ref{app:extrallm}.

\textbf{Observation 1: NAS yields the strongest long-horizon sequential edit performance.}
Aggregating results over \emph{both} CounterFact and ZsRE and all three backbones, and comparing each metric to the
\emph{strongest baseline under the same evaluation block}, NAS improves edit performance by \textbf{+11.5pp} (Eff.) and
\textbf{+9.9pp} (Gen.) on average, with specificity essentially unchanged (within 0.3pp on average). 
These gains are consistent with our hidden-representation drift diagnosis (Fig.~\ref{fig:pca_drift_insets}): NAS suppresses the target-layer OOD shift induced by vanilla L\&E, and improves long-horizon Eff./Gen.\ without a stability--strength trade-off despite constraining the update magnitude.
NAS also surpasses recent lifelong-focused editors such as \textsc{UltraEdit} and \textsc{RLEdit} on most metrics, often by a wide margin.
The advantage is most pronounced on GPT-J, where NAS boosts CounterFact efficacy from \textbf{78.1} to \textbf{98.7}
(\textbf{+20.6pp}) and generalization from \textbf{54.9} to \textbf{90.7} (\textbf{+35.8pp}) over the strongest 
baseline, while also improving specificity (\textbf{54.7} $\rightarrow$ \textbf{60.8}, \textbf{+6.0pp}).

\textbf{Observation 2: NAS maintains fluent generations and improves consistency.}
Beyond edit performance, NAS remains competitive on \textit{Fluency} while delivering substantially higher
\textit{Consistency} under long-horizon sequential editing. Aggregated across the three backbones, NAS improves fluency by
\textbf{+8.9} and consistency by \textbf{+13.5} on average compared to the best baseline within the same evaluation
block. The gains are most pronounced on Llama3-8B, where NAS raises fluency
from \textbf{596.2} to \textbf{619.7} (\textbf{+23.5}) and boosts consistency from \textbf{12.2} to \textbf{31.9} (\textbf{+19.7})
over the strongest baseline.

\subsection{General Capability Retention (RQ2)}
To test whether long-horizon sequential editing compromises a model's \emph{general capability}, we track F1 on
\emph{GLUE-style tasks} as edits accumulate. We perform sequential L\&E updates over the \emph{entire} CounterFact stream
(\counterfactStreamSize{} edits) for five representative editors (MEMIT, PRUNE, RECT, AlphaEdit, and NAS), and plot the capability
trajectories throughout the editing process (Fig.~\ref{fig:rq2_glue_f1_seq}; insets zoom into 0--4k edits). We observe consistent
trends on additional backbones (Qwen2.5 and GPT-J), with full results in Appendix~\ref{app:extrageneral}.

\textbf{Observation 1: NAS avoids catastrophic capability degradation.}
Across the full \counterfactStreamSize-edit stream, NAS maintains \emph{non-trivial} GLUE-style performance without exhibiting
catastrophic collapse. Most tasks remain stable over the entire horizon; MRPC shows a more noticeable decline, yet does
not degenerate into a near-zero regime.

\textbf{Observation 2: NAS extends the capability-preserving horizon by $\ge 5\times$.}
 The prior L\&E baselines degrade rapidly: their GLUE-style performance drops to near-zero within the first few
thousand edits (typically $\le$4k). In contrast, NAS extends the capability-preserving horizon from roughly $\sim$4k
edits to at least \counterfactStreamSize{} edits, corresponding to an improvement of \emph{at least} $5\times$.

\subsection{Plug-in Gains in Stability and Performance (RQ3)}
\label{sec:exp:rq3_plugin}

We evaluate NAS as a plug-in for L\&E editors by attaching NAS to four representative L\&E baselines with only a \emph{one-line} integration.
(MEMIT, PRUNE, RECT, and AlphaEdit) and run long-horizon sequential editing on CounterFact. Main results are reported
on Llama3-8B, and we replicate the same study on GPT-J in Appendix~\ref{app:extraplugin}. Unless otherwise stated, we
report editing quality at a shared reference point---the collapse point (CP) of each corresponding base editor.

\textbf{Observation 1: NAS suppresses norm explosion and delays collapse across base editors.}
Figure~\ref{fig:rq3_ts_and_bars}(a) shows dual-axis trajectories during sequential editing, where solid lines track the
post-edit score and dashed lines track the norm statistic $R_n$. Across all four editors, the base variants exhibit rapid
growth in $R_n$ followed by a sharp collapse in score. After integrating NAS, the explosive growth is suppressed and the
collapse point is substantially pushed back: $3.75\times$ for MEMIT and PRUNE (CP: 400 $\rightarrow$ 1.5k), $3.38\times$
for RECT (400 $\rightarrow$ 1.4k), and $>5.25\times$ for AlphaEdit (4k $\rightarrow$ $>$21k). This indicates that NAS acts as
a robust stabilization plug-in across heterogeneous L\&E update rules.

\textbf{Observation 2: NAS improves editing quality for each base editor.}
 NAS-augmented editors achieve substantially higher post-edit success on Eff./Gen./Spe. at the reference point, as shown in Fig.~\ref{fig:rq3_ts_and_bars}(b). Averaged across editors and metrics, NAS
improves success by \textbf{+37.4 percentage points} (51.9\% $\rightarrow$ 89.3\%), corresponding to a \textbf{+72.2\% relative}
increase. The gains are broad across efficacy, generalization, and specificity, suggesting that stabilization translates into
systematically improved editing behavior rather than isolated wins.

We provide plug-in general-capability results at the same reference point in Appendix~\ref{app:plugin_general_capability}.

\subsection{Single-Edit Plasticity under Norm Anchoring (RQ4)}

A natural concern is that NAS may improve long-horizon stability simply by weakening each individual edit. 
To test this, we evaluate independent single edits on 1,000 CounterFact examples, where each edit starts from the original base model and is evaluated in isolation. 
We compare each base editor with and without NAS under the same editing configuration. 
As shown in Table~\ref{tab:single_edit}, NAS does not systematically reduce single-edit efficacy.

We further stress-test the strongest intervention cases by selecting the top 25\% examples whose solved value vectors are most strongly downscaled by NAS. 
Even in this hard subset, AlphaEdit+NAS remains neutral to slightly better, while MEMIT+NAS shows only a small score drop. 
These results indicate that the long-horizon gains of NAS are not explained by a broad stability--plasticity trade-off.

\begin{table}[t]
\centering
\small
\caption{
Single-edit plasticity under NAS. Full absolute numbers are in Appendix~\ref{app:single_edit}.
}
\vspace{0.5mm}
\label{tab:single_edit}
\begin{tabular}{lccrrrr}
\toprule
Editor & Split & NAS effect & $\Delta$Eff. & $\Delta$Gen. & $\Delta$Spe. & $\Delta$Score \\
\midrule
MEMIT     & All       & On -- Off & $0.00$  & $+1.80$ & $-0.06$ & $+0.70$ \\
AlphaEdit & All       & On -- Off & $+1.00$ & $+2.90$ & $-0.18$ & $+1.33$ \\
\midrule
MEMIT     & Top-25\%  & On -- Off & $-0.80$ & $-2.80$ & $+0.16$ & $-1.17$ \\
AlphaEdit & Top-25\%  & On -- Off & $0.00$  & $0.00$  & $+0.64$ & $+0.24$ \\
\bottomrule
\end{tabular}
\vspace{-1.5mm}
\end{table}

Additional robustness and breadth checks, including anchor mis-specification, non-stationary edit orders, multi-hop counterfactual editing, and shorter KnowEdit streams, are provided in Appendix~\ref{app:c}.

\vspace{-1mm}
\section{Related Work}

Model editing methods can be broadly divided into \emph{in-weight} editors, which overwrite pretrained parameters, and \emph{backbone-frozen} editors, which store edits in auxiliary modules or external memory~\citep{Wangsurvey}. 
Non-L\&E in-weight methods include FT~\citep{ft2020}, MEND~\citep{mitchell2022mend}, UltraEdit~\citep{gu2025ultraedittrainingsubjectmemoryfree}, and RLEdit~\citep{li2025rledit}, which directly optimize, predict, or learn parameter updates without an explicit locate-and-edit step. 
Backbone-frozen methods such as GRACE~\citep{hartvigsen2023grace}, MELO~\citep{yu2024melo}, WISE~\citep{wang2024wise}, and MEMOIR~\citep{wang2025memoirlifelongmodelediting} avoid overwriting the backbone by routing edits through adapters or memory, but require additional inference-time components.

Our work focuses on \emph{Locate-and-Edit (L\&E)} methods, which perform localized updates at causal FFN sites. 
ROME rewrites individual facts with rank-one updates~\citep{meng2022rome}, while MEMIT extends this idea to mass editing with structured multi-layer updates~\citep{meng2023memit}. 
Recent sequential L\&E variants, including AlphaEdit, PRUNE, RECT, ENCORE, and LyapLock, introduce subspace constraints, perturbation control, relative-change restrictions, norm regularization, or long-run preservation objectives to reduce interference and delay degradation~\citep{fang2025alphaedit,ma2025prune,gu2024rect,gupta2025norm15k,wang2025lyaplockboundedknowledgepreservation}. 
NAS is complementary to these approaches: instead of changing the locate step, editor objective, or adding inference-time modules, it anchors the \emph{final solved value vector} written by each edit, directly targeting the residual norm-feedback dynamics that can persist under increment-level controls.

\vspace{-1mm}
\section{Conclusion, Limitation and Future Work}
We identify a positive norm-feedback loop in sequential L\&E editing and propose \textsc{NAS}, a vector-level norm-control method that breaks this loop by anchoring the solved value vector rather than only regularizing update increments. 
This simple intervention substantially improves editing lifespan and long-run performance with negligible overhead. 
A limitation is that NAS uses a fixed original-model reference norm, which may be suboptimal when edit difficulty, locality requirements, or the edit distribution changes substantially; future work should explore adaptive or conditional anchoring while preserving the plug-in simplicity of NAS.

\bibliographystyle{plainnat}
\bibliography{NAS}

@misc{geva2021kvm,
      title={Transformer Feed-Forward Layers Are Key-Value Memories}, 
      author={Mor Geva and Roei Schuster and Jonathan Berant and Omer Levy},
      year={2021},
      eprint={2012.14913},
      archivePrefix={arXiv},
      primaryClass={cs.CL},
      url={https://arxiv.org/abs/2012.14913}, 
}

@article{zhang2024comprehensive,
  title={A Comprehensive Study of Knowledge Editing for Large Language Models},
  author={Zhang, Ningyu and Yao, Yunzhi and Tian, Bozhong and Wang, Peng and Deng, Shumin and Wang, Mengru and Xi, Zekun and Mao, Shengyu and Zhang, Jintian and Ni, Yuansheng and Cheng, Siyuan and Xu, Ziwen and Xu, Xin and Gu, Jia-Chen and Jiang, Yong and Xie, Pengjun and Huang, Fei and Liang, Lei and Zhang, Zhiqiang and Zhu, Xiaowei and Zhou, Jun and Chen, Huajun},
  journal={arXiv preprint arXiv:2401.01286},
  year={2024}
}

@inproceedings{zhong-etal-2023-mquake,
  title = "{MQ}u{AKE}: Assessing Knowledge Editing in Language Models via Multi-Hop Questions",
  author = "Zhong, Zexuan and Wu, Zhengxuan and Manning, Christopher and Potts, Christopher and Chen, Danqi",
  booktitle = "Proceedings of the 2023 Conference on Empirical Methods in Natural Language Processing",
  month = dec,
  year = "2023",
  address = "Singapore",
  publisher = "Association for Computational Linguistics",
  url = "https://aclanthology.org/2023.emnlp-main.971/",
  doi = "10.18653/v1/2023.emnlp-main.971",
  pages = "15686--15702"
}

@inproceedings{yaosurvey,
    title = "Editing Large Language Models: Problems, Methods, and Opportunities",
    author = "Yao, Yunzhi  and
      Wang, Peng  and
      Tian, Bozhong  and
      Cheng, Siyuan  and
      Li, Zhoubo  and
      Deng, Shumin  and
      Chen, Huajun  and
      Zhang, Ningyu",
    booktitle = "Proceedings of the 2023 Conference on Empirical Methods in Natural Language Processing",
    month = dec,
    year = "2023",
    address = "Singapore",
    publisher = "Association for Computational Linguistics",
    url = "https://aclanthology.org/2023.emnlp-main.632/",
    doi = "10.18653/v1/2023.emnlp-main.632",
    pages = "10222--10240",
}

@misc{meng2022rome,
      title={Locating and Editing Factual Associations in GPT}, 
      author={Kevin Meng and David Bau and Alex Andonian and Yonatan Belinkov},
      year={2023},
      eprint={2202.05262},
      archivePrefix={arXiv},
      primaryClass={cs.CL},
      url={https://arxiv.org/abs/2202.05262}, 
}

@misc{meng2023memit,
      title={Mass-Editing Memory in a Transformer}, 
      author={Kevin Meng and Arnab Sen Sharma and Alex Andonian and Yonatan Belinkov and David Bau},
      year={2023},
      eprint={2210.07229},
      archivePrefix={arXiv},
      primaryClass={cs.CL},
      url={https://arxiv.org/abs/2210.07229}, 
}

@article{fang2025alphaedit,
  title={Alphaedit: Null-space constrained knowledge editing for language models},
  author={Fang, Junfeng and Jiang, Houcheng and Wang, Kun and Ma, Yunshan and Jie, Shi and Wang, Xiang and He, Xiangnan and Chua, Tat-Seng},
  journal={arXiv preprint arXiv:2410.02355},
  year={2024}
}

@article{ma2025prune,
  title={Perturbation-restrained sequential model editing},
  author={Ma, Jun-Yu and Wang, Hong and Xu, Hao-Xiang and Ling, Zhen-Hua and Gu, Jia-Chen},
  journal={arXiv preprint arXiv:2405.16821},
  year={2024}
}

@misc{gu2024rect,
      title={Model Editing Harms General Abilities of Large Language Models: Regularization to the Rescue}, 
      author={Jia-Chen Gu and Hao-Xiang Xu and Jun-Yu Ma and Pan Lu and Zhen-Hua Ling and Kai-Wei Chang and Nanyun Peng},
      year={2024},
      eprint={2401.04700},
      archivePrefix={arXiv},
      primaryClass={cs.CL},
      url={https://arxiv.org/abs/2401.04700}, 
}

@article{mitchell2022mend,
  author       = {Eric Mitchell and
                  Charles Lin and
                  Antoine Bosselut and
                  Chelsea Finn and
                  Christopher D. Manning},
  title        = {Fast Model Editing at Scale},
  journal      = {CoRR},
  volume       = {abs/2110.11309},
  year         = {2021},
  url          = {https://arxiv.org/abs/2110.11309},
  eprinttype    = {arXiv},
  eprint       = {2110.11309},
  bibsource    = {dblp computer science bibliography, https://dblp.org}
}

@misc{hartvigsen2023grace,
      title={Aging with GRACE: Lifelong Model Editing with Discrete Key-Value Adaptors}, 
      author={Thomas Hartvigsen and Swami Sankaranarayanan and Hamid Palangi and Yoon Kim and Marzyeh Ghassemi},
      year={2023},
      eprint={2211.11031},
      archivePrefix={arXiv},
      primaryClass={cs.LG},
      url={https://arxiv.org/abs/2211.11031}, 
}

@misc{yu2024melo,
      title={MELO: Enhancing Model Editing with Neuron-Indexed Dynamic LoRA}, 
      author={Lang Yu and Qin Chen and Jie Zhou and Liang He},
      year={2023},
      eprint={2312.11795},
      archivePrefix={arXiv},
      primaryClass={cs.CL},
      url={https://arxiv.org/abs/2312.11795}, 
}

@misc{wang2024wise,
      title={WISE: Rethinking the Knowledge Memory for Lifelong Model Editing of Large Language Models}, 
      author={Peng Wang and Zexi Li and Ningyu Zhang and Ziwen Xu and Yunzhi Yao and Yong Jiang and Pengjun Xie and Fei Huang and Huajun Chen},
      year={2024},
      eprint={2405.14768},
      archivePrefix={arXiv},
      primaryClass={cs.CL},
      url={https://arxiv.org/abs/2405.14768}, 
}

@misc{Wangsurvey,
      title={Knowledge Editing for Large Language Models: A Survey}, 
      author={Song Wang and Yaochen Zhu and Haochen Liu and Zaiyi Zheng and Chen Chen and Jundong Li},
      year={2024},
      eprint={2310.16218},
      archivePrefix={arXiv},
      primaryClass={cs.CL},
      url={https://arxiv.org/abs/2310.16218}, 
}

@inproceedings{petroni-etal-2019-language,
    title = "Language Models as Knowledge Bases?",
    author = {Petroni, Fabio  and
      Rockt{\"a}schel, Tim  and
      Riedel, Sebastian  and
      Lewis, Patrick  and
      Bakhtin, Anton  and
      Wu, Yuxiang  and
      Miller, Alexander},
    booktitle = "Proceedings of the 2019 Conference on Empirical Methods in Natural Language Processing and the 9th International Joint Conference on Natural Language Processing (EMNLP-IJCNLP)",
    month = nov,
    year = "2019",
    address = "Hong Kong, China",
    publisher = "Association for Computational Linguistics",
    url = "https://aclanthology.org/D19-1250/",
    doi = "10.18653/v1/D19-1250",
    pages = "2463--2473",
}

@inproceedings{roberts-etal-2020-much,
    title = "How Much Knowledge Can You Pack Into the Parameters of a Language Model?",
    author = "Roberts, Adam  and
      Raffel, Colin  and
      Shazeer, Noam",
    booktitle = "Proceedings of the 2020 Conference on Empirical Methods in Natural Language Processing (EMNLP)",
    month = nov,
    year = "2020",
    address = "Online",
    publisher = "Association for Computational Linguistics",
    url = "https://aclanthology.org/2020.emnlp-main.437/",
    doi = "10.18653/v1/2020.emnlp-main.437",
    pages = "5418--5426",
}

@inproceedings{lin-etal-2022-truthfulqa,
    title = "{T}ruthful{QA}: Measuring How Models Mimic Human Falsehoods",
    author = "Lin, Stephanie  and
      Hilton, Jacob  and
      Evans, Owain",
    booktitle = "Proceedings of the 60th Annual Meeting of the Association for Computational Linguistics (Volume 1: Long Papers)",
    month = may,
    year = "2022",
    address = "Dublin, Ireland",
    publisher = "Association for Computational Linguistics",
    url = "https://aclanthology.org/2022.acl-long.229/",
    doi = "10.18653/v1/2022.acl-long.229",
    pages = "3214--3252",
}

@inproceedings{de-cao-etal-2021-editing,
    title = "Editing Factual Knowledge in Language Models",
    author = "De Cao, Nicola  and
      Aziz, Wilker  and
      Titov, Ivan",
    booktitle = "Proceedings of the 2021 Conference on Empirical Methods in Natural Language Processing",
    month = nov,
    year = "2021",
    address = "Online and Punta Cana, Dominican Republic",
    publisher = "Association for Computational Linguistics",
    url = "https://aclanthology.org/2021.emnlp-main.522/",
    doi = "10.18653/v1/2021.emnlp-main.522",
    pages = "6491--6506",
}

@article{dhingra-etal-2022-time,
    title = "Time-Aware Language Models as Temporal Knowledge Bases",
    author = "Dhingra, Bhuwan  and
      Cole, Jeremy R.  and
      Eisenschlos, Julian Martin  and
      Gillick, Daniel  and
      Eisenstein, Jacob  and
      Cohen, William W.",
    journal = "Transactions of the Association for Computational Linguistics",
    volume = "10",
    year = "2022",
    address = "Cambridge, MA",
    publisher = "MIT Press",
    url = "https://aclanthology.org/2022.tacl-1.15/",
    doi = "10.1162/tacl_a_00459",
    pages = "257--273",
}

@inproceedings{
zhang2024dissecting,
title={Dissecting learning and forgetting in language model finetuning},
author={Xiao Zhang and Ji Wu},
booktitle={The Twelfth International Conference on Learning Representations},
year={2024},
url={https://openreview.net/forum?id=tmsqb6WpLz}
}

@inproceedings{Gupta2024ModelEA,
  title={Model Editing at Scale leads to Gradual and Catastrophic Forgetting},
  author={Akshat Gupta and Anurag Rao and Gopala Krishna Anumanchipalli},
  booktitle={Annual Meeting of the Association for Computational Linguistics},
  year={2024},
  url={https://api.semanticscholar.org/CorpusID:266999650}
}

@misc{zhang2025queueeditstructuralselfcorrectionsequential,
      title={QueueEDIT: Structural Self-Correction for Sequential Model Editing in LLMs}, 
      author={Taolin Zhang and Haidong Kang and Dongyang Li and Qizhou Chen and Chengyu Wang Xiaofeng He and Richang Hong},
      year={2025},
      eprint={2506.17864},
      archivePrefix={arXiv},
      primaryClass={cs.CL},
      url={https://arxiv.org/abs/2506.17864}, 
}

@article{zsrelevy,
  author       = {Omer Levy and
                  Minjoon Seo and
                  Eunsol Choi and
                  Luke Zettlemoyer},
  title        = {Zero-Shot Relation Extraction via Reading Comprehension},
  journal      = {CoRR},
  volume       = {abs/1706.04115},
  year         = {2017},
  url          = {http://arxiv.org/abs/1706.04115},
  eprinttype    = {arXiv},
  eprint       = {1706.04115},
  bibsource    = {dblp computer science bibliography, https://dblp.org}
}

@misc{llama3,
      title = {The {Llama 3} Herd of Models},
      author = {Grattafiori, Aaron and others},
      note = {AI@Meta},
      year = {2024},
      eprint = {2407.21783},
      archivePrefix = {arXiv},
      primaryClass = {cs.AI},
      url = {https://arxiv.org/abs/2407.21783},
}

@misc{qwen2025qwen25technicalreport,
      title={Qwen2.5 Technical Report}, 
      author={Qwen and : and An Yang and Baosong Yang and Beichen Zhang and Binyuan Hui and Bo Zheng and Bowen Yu and Chengyuan Li and Dayiheng Liu and Fei Huang and Haoran Wei and Huan Lin and Jian Yang and Jianhong Tu and Jianwei Zhang and Jianxin Yang and Jiaxi Yang and Jingren Zhou and Junyang Lin and Kai Dang and Keming Lu and Keqin Bao and Kexin Yang and Le Yu and Mei Li and Mingfeng Xue and Pei Zhang and Qin Zhu and Rui Men and Runji Lin and Tianhao Li and Tianyi Tang and Tingyu Xia and Xingzhang Ren and Xuancheng Ren and Yang Fan and Yang Su and Yichang Zhang and Yu Wan and Yuqiong Liu and Zeyu Cui and Zhenru Zhang and Zihan Qiu},
      year={2025},
      eprint={2412.15115},
      archivePrefix={arXiv},
      primaryClass={cs.CL},
      url={https://arxiv.org/abs/2412.15115}, 
}

@misc{gpt-j,
  author = {Wang, Ben and Komatsuzaki, Aran},
  title = {{GPT-J-6B: A 6 Billion Parameter Autoregressive Language Model}},
  howpublished = {\url{https://github.com/kingoflolz/mesh-transformer-jax}},
  year = 2021,
  month = May
}

@misc{gpt2-xl,
  title={Language Models are Unsupervised Multitask Learners},
  author={Alec Radford and Jeff Wu and Rewon Child and David Luan and Dario Amodei and Ilya Sutskever},
  year={2019},
  url={https://api.semanticscholar.org/CorpusID:160025533}
}

@article{ft2020,
  author       = {Chen Zhu and
                  Ankit Singh Rawat and
                  Manzil Zaheer and
                  Srinadh Bhojanapalli and
                  Daliang Li and
                  Felix X. Yu and
                  Sanjiv Kumar},
  title        = {Modifying Memories in Transformer Models},
  journal      = {CoRR},
  volume       = {abs/2012.00363},
  year         = {2020},
  url          = {https://arxiv.org/abs/2012.00363},
  eprinttype    = {arXiv},
  eprint       = {2012.00363},
  bibsource    = {dblp computer science bibliography, https://dblp.org}
}

@misc{glue,
      title={GLUE: A Multi-Task Benchmark and Analysis Platform for Natural Language Understanding}, 
      author={Alex Wang and Amanpreet Singh and Julian Michael and Felix Hill and Omer Levy and Samuel R. Bowman},
      year={2019},
      eprint={1804.07461},
      archivePrefix={arXiv},
      primaryClass={cs.CL},
      url={https://arxiv.org/abs/1804.07461}, 
}

@inproceedings{sst,
    title = "Recursive Deep Models for Semantic Compositionality Over a Sentiment Treebank",
    author = "Socher, Richard  and
      Perelygin, Alex  and
      Wu, Jean  and
      Chuang, Jason  and
      Manning, Christopher D.  and
      Ng, Andrew  and
      Potts, Christopher",
    booktitle = "Proceedings of the 2013 Conference on Empirical Methods in Natural Language Processing",
    month = oct,
    year = "2013",
    address = "Seattle, Washington, USA",
    publisher = "Association for Computational Linguistics",
    url = "https://aclanthology.org/D13-1170/",
    pages = "1631--1642"
}

@inproceedings{mrpc,
    title = "Automatically Constructing a Corpus of Sentential Paraphrases",
    author = "Dolan, William B.  and
      Brockett, Chris",
    booktitle = "Proceedings of the Third International Workshop on Paraphrasing ({IWP}2005)",
    year = "2005",
    url = "https://aclanthology.org/I05-5002/"
}

@inproceedings{rte,
  author       = {Luisa Bentivogli and
                  Bernardo Magnini and
                  Ido Dagan and
                  Hoa Trang Dang and
                  Danilo Giampiccolo},
  title        = {The Fifth {PASCAL} Recognizing Textual Entailment Challenge},
  booktitle    = {Proceedings of the Second Text Analysis Conference, {TAC} 2009, Gaithersburg,
                  Maryland, USA, November 16-17, 2009},
  publisher    = {{NIST}},
  year         = {2009},
  url          = {https://tac.nist.gov/publications/2009/additional.papers/RTE5\_overview.proceedings.pdf},
  bibsource    = {dblp computer science bibliography, https://dblp.org}
}

@article{cola,
    title = "Neural Network Acceptability Judgments",
    author = "Warstadt, Alex  and
      Singh, Amanpreet  and
      Bowman, Samuel R.",
    journal = "Transactions of the Association for Computational Linguistics",
    volume = "7",
    year = "2019",
    address = "Cambridge, MA",
    publisher = "MIT Press",
    url = "https://aclanthology.org/Q19-1040/",
    doi = "10.1162/tacl_a_00290",
    pages = "625--641",
}

@inproceedings{mnli,
    title = "A Broad-Coverage Challenge Corpus for Sentence Understanding through Inference",
    author = "Williams, Adina  and
      Nangia, Nikita  and
      Bowman, Samuel",
    booktitle = "Proceedings of the 2018 Conference of the North {A}merican Chapter of the Association for Computational Linguistics: Human Language Technologies, Volume 1 (Long Papers)",
    month = jun,
    year = "2018",
    address = "New Orleans, Louisiana",
    publisher = "Association for Computational Linguistics",
    url = "https://aclanthology.org/N18-1101/",
    doi = "10.18653/v1/N18-1101",
    pages = "1112--1122",
}

@misc{mmlu,
      title={Measuring Massive Multitask Language Understanding}, 
      author={Dan Hendrycks and Collin Burns and Steven Basart and Andy Zou and Mantas Mazeika and Dawn Song and Jacob Steinhardt},
      year={2021},
      eprint={2009.03300},
      archivePrefix={arXiv},
      primaryClass={cs.CY},
      url={https://arxiv.org/abs/2009.03300}, 
}

@misc{gu2025ultraedittrainingsubjectmemoryfree,
      title={UltraEdit: Training-, Subject-, and Memory-Free Lifelong Editing in Language Models}, 
      author={Xiaojie Gu and Ziying Huang and Jia-Chen Gu and Kai Zhang},
      year={2025},
      eprint={2505.14679},
      archivePrefix={arXiv},
      primaryClass={cs.CL},
      url={https://arxiv.org/abs/2505.14679}, 
}

@misc{li2025rledit,
      title={Reinforced Lifelong Editing for Language Models}, 
      author={Zherui Li and Houcheng Jiang and Hao Chen and Baolong Bi and Zhenhong Zhou and Fei Sun and Junfeng Fang and Xiang Wang},
      year={2025},
      eprint={2502.05759},
      archivePrefix={arXiv},
      primaryClass={cs.CL},
      url={https://arxiv.org/abs/2502.05759}, 
}

@misc{gupta2025norm15k,
      title={Lifelong Knowledge Editing requires Better Regularization}, 
      author={Akshat Gupta and Phudish Prateepamornkul and Maochuan Lu and Ahmed Alaa and Thomas Hartvigsen and Gopala Anumanchipalli},
      year={2025},
      eprint={2502.01636},
      archivePrefix={arXiv},
      primaryClass={cs.CL},
      url={https://arxiv.org/abs/2502.01636}, 
}

@misc{wang2025lyaplockboundedknowledgepreservation,
      title={LyapLock: Bounded Knowledge Preservation in Sequential Large Language Model Editing}, 
      author={Peng Wang and Biyu Zhou and Xuehai Tang and Jizhong Han and Songlin Hu},
      year={2025},
      eprint={2505.15702},
      archivePrefix={arXiv},
      primaryClass={cs.CL},
      url={https://arxiv.org/abs/2505.15702}, 
}

@misc{wang2025memoirlifelongmodelediting,
      title={MEMOIR: Lifelong Model Editing with Minimal Overwrite and Informed Retention for LLMs}, 
      author={Ke Wang and Yiming Qin and Nikolaos Dimitriadis and Alessandro Favero and Pascal Frossard},
      year={2025},
      eprint={2506.07899},
      archivePrefix={arXiv},
      primaryClass={cs.CL},
      url={https://arxiv.org/abs/2506.07899}, 
}

@inproceedings{thede2025wikibigedit,
  title={{WikiBigEdit: Understanding the Limits of Lifelong Knowledge Editing in LLMs}},
  author={Lukas Thede and Karsten Roth and Matthias Bethge and Zeynep Akata and Tom Hartvigsen},
  booktitle={Proceedings of the 42nd International Conference on Machine Learning (ICML)},
  year={2025}
}

\newpage
\appendix
\section{Experimental Setup and Implement Details}
\label{apA}

In this section, we describe the datasets and evaluation metrics used in our experiments, and detail the base-model capability evaluation, experimental setup, and baseline methods.

\subsection{Datasets}
\label{app:datasets}

\begin{itemize}
    \item \textbf{CounterFact.}
    CounterFact~\citep{meng2022rome} is a benchmark designed specifically for \emph{factual knowledge editing}. Each example specifies a factual triple in the form of a \emph{subject} and \emph{relation} (e.g., ``\emph{Paris} is the capital of \underline{France}'') together with a \emph{counterfactual target object} to be written into the model (e.g., replacing \underline{France} with a new object). The benchmark provides multiple prompt variants to query the edited knowledge (e.g., a direct rewrite prompt and paraphrases), as well as semantically related ``neighborhood'' prompts used to check locality/specificity. It is therefore commonly used to measure whether an editor succeeds on the intended rewrite while preserving behavior on nearby but non-target facts.

    \item \textbf{ZsRE.}
    ZsRE~\citep{zsrelevy} is a widely used factual probing dataset derived from relation extraction in a \emph{question-answering} format. Each instance corresponds to a (subject, relation) query rendered as a natural-language question (often with paraphrased variants), and the model is expected to generate the correct answer entity. In model editing, ZsRE is typically used to evaluate whether an edit can reliably update the answer to a targeted question while generalizing across question paraphrases, making it a complementary benchmark to CounterFact for assessing efficacy and generalization under diverse query formulations.

    \item \textbf{WikiBigEdit.}
WikiBigEdit~\citep{thede2025wikibigedit} is a large-scale \emph{lifelong} knowledge-editing benchmark constructed from real-world \emph{Wikidata snapshot diffs}, yielding a chronological stream of factual changes with over 500K question--answer pairs. Each example is grounded in a Wikidata (subject, relation, object) triplet and provides an \emph{update} question plus rephrases/persona variants to test generalization; it further includes dedicated \emph{locality} prompts (with expected answers) and optional \emph{multi-hop} questions to probe whether edits preserve unrelated knowledge and reasoning behavior. %
\end{itemize}

\subsection{Metrics}
\label{app:metrics}

We report standard knowledge-editing metrics on \textsc{ZsRE} and \textsc{CounterFact}, and additionally
introduce a long-horizon stability indicator based on the CounterFact score.
We also report \textsc{WikiBigEdit} metrics following the official benchmark protocol, namely \emph{Edit Success} (ES),
\emph{Generalization Success} (GS), and \emph{Locality Success} (LS).
For each edit instance $i$, let $f_\theta$ denote the (edited) language model, $(s_i, r_i)$ the edit
prompt constructed from the subject--relation pair, and $o_i$ the desired post-edit target output.
We use $N((s_i,r_i))$ to denote a set of paraphrased (semantically equivalent) prompts, and
$O((s_i,r_i))$ to denote a set of locality prompts used to probe non-target behavior.
In sequential-editing experiments, metrics are evaluated at a set of discrete checkpoints (shared across
all methods for a fixed base model), which enables us to define the \emph{collapse point} (CP) as the first
checkpoint where the CounterFact score falls below a predefined threshold.
Our definitions largely follow common practice in prior work (e.g., \citep{meng2022rome,meng2023memit,zhang2025queueeditstructuralselfcorrectionsequential,fang2025alphaedit}),
but we restate them here for completeness.

\subsubsection{ZsRE Metrics}
\label{app:metrics:zsre}

\paragraph{Efficacy.}
On \textsc{ZsRE}, efficacy is measured as the average top-1 accuracy on the edited questions:
\begin{equation}
\mathbb{E}_i\left\{\, o_i = \arg\max_{o}\,\mathbb{P}_{f_\theta}\!\left(o \mid (s_i,r_i)\right)\right\}.
\label{eq:zsre_eff}
\end{equation}

\paragraph{Generalization.}
Generalization evaluates whether the edit transfers to paraphrased queries. We compute the average
top-1 accuracy over the rephrased prompt set $N((s_i,r_i))$:
\begin{equation}
\mathbb{E}_i\left\{\, o_i = \arg\max_{o}\,\mathbb{P}_{f_\theta}\!\left(o \mid N((s_i,r_i))\right)\right\}.
\label{eq:zsre_gen}
\end{equation}

\paragraph{Specificity (Locality).}
Specificity measures whether predictions on locality prompts remain correct with respect to their dataset-provided locality answers $o_i^c$.
We evaluate this by checking whether the model's top-1 prediction on $O((s_i,r_i))$ matches the
(original) pre-edit answer $o_i^{c}$:
\begin{equation}
\mathbb{E}_i\left\{\, o_i^{c} = \arg\max_{o}\,\mathbb{P}_{f_\theta}\!\left(o \mid O((s_i,r_i))\right)\right\}.
\label{eq:zsre_spe}
\end{equation}

\subsubsection{CounterFact Metrics}
\label{app:metrics:counterfact}

On \textsc{CounterFact}, the evaluation is likelihood-based. Let $o_c^{i}$ be the model's original
(pre-edit) output for the edit prompt, and $o_i$ the desired new object to be written. We then define:

\paragraph{Efficacy (Rewrite Success).}
We count an edit as successful on the rewrite prompt if the new target is assigned higher probability
than the original output:
\begin{equation}
\mathbb{E}_i\left[\mathbb{P}_{f_\theta}\!\left[o_i \mid (s_i,r_i)\right] >
                 \mathbb{P}_{f_\theta}\!\left[o_c^{i} \mid (s_i,r_i)\right]\right].
\label{eq:cf_eff}
\end{equation}

\paragraph{Generalization (Paraphrase Success).}
We analogously evaluate paraphrase generalization on the rephrased prompt set $N((s_i,r_i))$:
\begin{equation}
\mathbb{E}_i\left[\mathbb{P}_{f_\theta}\!\left[o_i \mid N((s_i,r_i))\right] >
                 \mathbb{P}_{f_\theta}\!\left[o_c^{i} \mid N((s_i,r_i))\right]\right].
\label{eq:cf_gen}
\end{equation}

\paragraph{Specificity (Neighborhood/Locality Success).}
Specificity is computed on neighborhood prompts $O((s_i,r_i))$ (distinct but semantically related
subjects) by requiring the edited model to favor the correct/original completion over the edited target:
\begin{equation}
\mathbb{E}_i\left[\mathbb{P}_{f_\theta}\!\left[o_i \mid O((s_i,r_i))\right] <
                 \mathbb{P}_{f_\theta}\!\left[o_c^{i} \mid O((s_i,r_i))\right]\right].
\label{eq:cf_spe}
\end{equation}

\paragraph{Score (Harmonic Mean).}
In addition to reporting \textit{Efficacy}, \textit{Generalization}, and \textit{Specificity} separately,
we also summarize CounterFact performance with a single \emph{overall score} defined as their harmonic mean:
\begin{equation}
\mathrm{Score}
\;\coloneqq\;
\mathrm{H}\!\left(\mathrm{Eff.},\,\mathrm{Gen.},\,\mathrm{Spe.}\right)
\;=\;
\frac{3}{\frac{1}{\mathrm{Eff.}}+\frac{1}{\mathrm{Gen.}}+\frac{1}{\mathrm{Spe.}}}.
\label{eq:cf_score_hmean}
\end{equation}
Here, $\mathrm{Eff.}$, $\mathrm{Gen.}$, and $\mathrm{Spe.}$ are the dataset-level success rates defined above
(Eqs.~\ref{eq:cf_eff}--\ref{eq:cf_spe}). 

\paragraph{Collapse Point (CP@60).}
Let $\mathcal{T}_m=\{t_1,t_2,\ldots\}$ be the evaluation checkpoints for base model $m$, and let
$\mathrm{Score}(t)$ denote the CounterFact score at checkpoint $t$.
We define the collapse point as the earliest checkpoint where the score falls below a threshold:
\begin{equation}
\mathrm{CP}@60
\;\coloneqq\;
\min\left\{\, t \in \mathcal{T}_m\;:\;\mathrm{Score}(t)\le 60 \right\}.
\label{eq:cp_def}
\end{equation}
Due to discrete evaluation, the true collapse (if any) occurs within the interval
$(t^{-}, \mathrm{CP}@60]$, where $t^{-}$ is the immediate predecessor of $\mathrm{CP}@60$ in $\mathcal{T}_m$.

\paragraph{Fluency (Generation Entropy).}
To detect degenerate generations with excessive repetition, we compute an $n$-gram entropy score:
\begin{equation}
-\frac{2}{3}\sum_{k} g_{2}(k)\log_{2} g_{2}(k) \;-\; \frac{4}{3}\sum_{k} g_{3}(k)\log_{2} g_{3}(k),
\label{eq:cf_flu}
\end{equation}
where $g_n(\cdot)$ denotes the empirical frequency distribution of $n$-grams in the model output.

\paragraph{Consistency (Reference Score).}
Consistency evaluates whether the model's generated description remains aligned with a reference text.
Given a subject $s$, we compute the cosine similarity between the TF-IDF vector of the model-generated
text and that of a reference Wikipedia passage about the corresponding object.

\subsubsection{WikiBigEdit Metrics}
\label{app:metrics:wikibigedit}

On \textsc{WikiBigEdit}, the evaluation is accuracy-based on question--answer prompts.
Let $(s_i,r_i)$ denote the update query for edit instance $i$, and $o_i$ the desired post-edit target output. We use $N((s_i,r_i))$
to denote a set of generalization queries (e.g., rephrases/persona variants), and $O((s_i,r_i))$ to denote a set of
locality probes paired with edit $i$, whose answers should remain unchanged (denoted by $o_i^{c}$).

\paragraph{ES (Edit Success).}
We measure edit success as the average top-1 accuracy on the update queries:
\begin{equation}
\mathbb{E}_i\left\{\, o_i = \arg\max_{o}\,\mathbb{P}_{f_\theta}\!\left(o \mid (s_i,r_i)\right)\right\}.
\label{eq:wbe_es}
\end{equation}

\paragraph{GS (Generalization Success).}
Generalization evaluates whether the edit transfers to the associated generalization set $N((s_i,r_i))$:
\begin{equation}
\mathbb{E}_i\left\{\, o_i = \arg\max_{o}\,\mathbb{P}_{f_\theta}\!\left(o \mid N((s_i,r_i))\right)\right\}.
\label{eq:wbe_gs}
\end{equation}

\paragraph{LS (Locality Success).}
Locality evaluates whether predictions on paired locality probes $O((s_i,r_i))$ match their original (pre-edit) answers $o_i^{c}$:
\begin{equation}
\mathbb{E}_i\left\{\, o_i^{c} = \arg\max_{o}\,\mathbb{P}_{f_\theta}\!\left(o \mid O((s_i,r_i))\right)\right\}.
\label{eq:wbe_ls}
\end{equation}

\subsection{Implementation details}
\label{app:implementNAS}

\begin{table}[H]
\centering
\small
\renewcommand{\arraystretch}{1.06}
\setlength{\tabcolsep}{6pt}
\caption{
Ablation on hyperparameter $N$. We report runtime (seconds)
and the raw-mean estimate of $\lVert v^\mathrm{new} \rVert$.
}
\label{tab:n_ablation_raw}
\begin{tabular}{l r c c}
\toprule
\textbf{Model} & \textbf{$N$} & \textbf{Time (s)} & \textbf{Raw $\lVert v^\mathrm{new} \rVert$} \\
\midrule
\multirow{5}{*}{Llama3-8B}
& 100  & $47.15 \pm 5.68$   & $5.298 \pm 0.040$ \\
& 300  & $143.33 \pm 1.95$  & $5.309 \pm 0.019$ \\
& 500  & $239.13 \pm 4.05$  & $5.320 \pm 0.020$ \\
& 1000 & $469.34 \pm 6.81$  & $5.325 \pm 0.015$ \\
& 2000 & $936.79 \pm 8.51$  & $5.322 \pm 0.007$ \\
\midrule
\multirow{5}{*}{GPT-J-6B}
& 100  & $13.07 \pm 1.57$   & $63.248 \pm 1.102$ \\
& 300  & $41.43 \pm 1.37$   & $62.872 \pm 0.634$ \\
& 500  & $67.84 \pm 2.54$   & $62.946 \pm 0.561$ \\
& 1000 & $135.06 \pm 1.78$  & $63.075 \pm 0.412$ \\
& 2000 & $268.83 \pm 2.99$  & $63.038 \pm 0.179$ \\
\bottomrule
\end{tabular}
\end{table}

\paragraph{Estimating the anchor norm.}
NAS requires a single scalar anchor norm $\tau \coloneqq \bigl(\mathbb{E}\| v^{\mathrm{new}}\|^2\bigr)^{1/2}$ measured on an original (unedited)
model. In all experiments, we estimate this expectation by performing pilot edits on $N=1000$ randomly sampled facts and
computing the empirical mean of the resulting $\| v^{\mathrm{new}}\|^2$ values. We use the same editing site and the same
base editor configuration as in the main sequential-editing runs.

\paragraph{Integration with base editors.}
NAS is implemented as a lightweight post-processing step on the target value vector produced by the base editor.
When inserting NAS into an editor, we only introduce two NAS-specific hyperparameters:
(i) the pilot sample size $N$ (fixed to $1000$ throughout this paper), and
(ii) a boolean flag \texttt{use\_nas} indicating whether to enable the rescaling rule.
All other hyperparameters and logic---including the computation of $\tilde v_n^{\mathrm{new}}$, optimization settings, and update rules---are
inherited \emph{unchanged} from the attached editor. Unless otherwise stated, when we refer to ``NAS'' as an independent editing method,
it denotes \textbf{AlphaEdit} equipped with NAS (i.e., AlphaEdit\,+\,NAS).

\paragraph{Sensitivity to $N$.}
We ablate the hyperparameter $N$ over $\{100,300,500,1000,2000\}$ with 5 restarts.
Across both Llama3-8B and GPT-J-6B, the raw-mean estimate of $\lVert v^\mathrm{new} \rVert$ remains highly stable as $N$ increases
(e.g., Llama3-8B varies only from $5.298$ at $N{=}100$ to $5.322$ at $N{=}2000$, and GPT-J-6B stays within $62.87$--$63.25$),
with variability across restarts generally shrinking for larger $N$.
In contrast, the runtime increases approximately linearly with $N$ (e.g., on Llama3-8B from $47.15$s at $N{=}100$ to $936.79$s at $N{=}2000$),
indicating that larger $N$ primarily trades compute for marginally reduced estimator variance rather than changing the central tendency.

\paragraph{Compute resources.}
All experiments were conducted on a single NVIDIA H100 GPU. For in-weight Locate-and-Edit (L\&E) editors (e.g.,
ROME/MEMIT/PRUNE/RECT/AlphaEdit), running sequential editing on our primary base models typically requires at
least 48\,GB of GPU memory due to the model size and the per-edit optimization/update procedure. Importantly, enabling
NAS does not introduce additional GPU-memory overhead.

\subsection{Baselines}\label{app:baselines} 
All baselines are evaluated using the \textbf{official implementations and recommended hyperparameters released by the authors}. The only exception is PRUNE, for which we use the implementation provided in the \textbf{AlphaEdit} codebase \citep{fang2025alphaedit}.

\begin{itemize}
\item \textbf{FT (Fine-Tuning). \citep{ft2020}} A straightforward baseline that directly fine-tunes the model on the edited fact via gradient-based optimization. While often effective at rewriting the target behavior, standard fine-tuning can introduce broader parameter changes and thus may cause unintended side effects on unrelated knowledge. 
\item \textbf{MEMIT. \citep{meng2023memit}} A scalable in-weight editing method designed for efficiently injecting many factual updates. It generalizes single-edit rank-constrained updates to multi-edit settings by composing a structured weight update (often distributed across layers), enabling large batches or long streams of edits with improved efficiency. 
\item \textbf{PRUNE. \citep{ma2025prune}} A regularization-based editor that constrains edit-induced perturbations to preserve model fidelity under sequential updates. It controls the geometry (e.g., update conditioning) of the applied change to suppress overly aggressive directions, thereby mitigating degradation and side effects over long edit horizons. 
\item \textbf{RECT. \citep{gu2024rect}} An editing method that restricts the relative change of weights during an update, typically by sparsifying or truncating the weight delta so that only the most relevant subset of parameters is modified. This strategy retains the base model's general capabilities while still achieving high edit efficacy on the target knowledge. 
\item \textbf{AlphaEdit. \citep{fang2025alphaedit}} A constrained in-weight editor that shapes the update to reduce interference with pre-existing knowledge. It enforces an approximate invariance constraint (e.g., a null-space projection) so that the update primarily affects the target association while minimizing unintended changes to other facts. 
\item \textbf{UltraEdit. \citep{gu2025ultraedittrainingsubjectmemoryfree}} A training- and memory-free editing method that computes the weight update in a single step from a hidden state and its gradient, without any additional fine-tuning. UltraEdit also employs a continual feature normalization strategy to adapt to distribution shifts across sequential edits, enabling ultra-fast and scalable knowledge updates with minimal interference. 
\item \textbf{RLEdit. \citep{li2025rledit}} A reinforcement learning-based editor that formulates sequential model editing as a decision process for a hypernetwork. By treating editing success as a reward and optimizing the hypernetwork over entire edit sequences, RLEdit adapts to dynamic model changes and achieves superior editing efficacy and efficiency in lifelong settings. 
\item \textbf{GRACE. \citep{hartvigsen2023grace}} A lifelong editing approach that stores new knowledge in auxiliary key--value style adapters rather than modifying backbone parameters. Each edit is written into a dedicated latent adapter, and a retrieval mechanism activates the corresponding adapter for relevant inputs, supporting a large number of edits with reduced interference on original knowledge. 
\item \textbf{MELO. \citep{yu2024melo}} A plug-in editor that realizes edits via dynamically activated low-rank adapters (e.g., LoRA-style modules). New facts are stored in separate adapter modules that are selectively invoked for matching inputs, providing modularity and strong isolation of edits without changing the core model weights. 
\item \textbf{WISE. \citep{wang2024wise}} A dual-memory approach that separates original knowledge retention from edited knowledge storage. It writes updates into an auxiliary ``side'' memory and uses a routing mechanism to combine or choose between the base model and the edited memory at inference time, aiming to reduce interference and improve stability under continual updates. 

\item \textbf{LyapLock. \citep{wang2025lyaplockboundedknowledgepreservation}} A sequential L\&E wrapper that formulates editing as a long-term constrained optimization problem and uses Lyapunov optimization (virtual queues) to enforce cumulative knowledge-preservation constraints while remaining compatible with existing editors.

\item \textbf{ENCORE. \citep{gupta2025norm15k}} A regularization-oriented L\&E extension that combines Most-Probable Early Stopping (MPES) with Frobenius-norm regularization on MEMIT-style closed-form updates to mitigate over-optimization and norm growth in long edit streams.

\item \textbf{MEMOIR. \citep{wang2025memoirlifelongmodelediting}} A backbone-frozen lifelong editor that writes edits into a residual memory module via sparse, sample-dependent masked updates (TopHash) and selectively retrieves relevant memory at inference to reduce interference.
\end{itemize}

\subsection{General Capability Evaluation}
\label{app:general_capability}

Beyond edit-centric metrics, we also monitor whether long streams of edits erode the base model's
\emph{general-purpose} language understanding and reasoning ability.
This type of collateral degradation under (lifelong) sequential editing has been repeatedly observed in
recent evaluations of model editing at scale, motivating routine reporting of downstream benchmark
performance alongside editing success \citep{fang2025alphaedit,zhang2025queueeditstructuralselfcorrectionsequential}.
Following this common practice, we evaluate the edited model on five representative tasks from
\textsc{GLUE} \citep{glue} and on \textsc{MMLU} \citep{mmlu} at a set of checkpoints
during the edit stream.


\begin{itemize}
    \item \textbf{SST-2 (SST).}
    A sentence-level sentiment classification task derived from movie reviews.
    The model predicts whether a review expresses \emph{positive} or \emph{negative} sentiment, probing
    basic lexical and compositional understanding as well as robustness to stylistic variation.

    \item \textbf{MRPC.}
    The Microsoft Research Paraphrase Corpus is a sentence-pair task that asks whether two sentences are
    paraphrases (i.e., semantically equivalent).
    It serves as a compact probe of semantic similarity judgments and sensitivity to small meaning changes.

    \item \textbf{RTE.}
    Recognizing Textual Entailment is a binary natural language inference (NLI) task.
    Given a premise and a hypothesis, the model predicts whether the hypothesis is entailed by the premise,
    providing a probe of logical/semantic inference under limited supervision.

    \item \textbf{CoLA.}
    The Corpus of Linguistic Acceptability asks whether a sentence is grammatically acceptable.
    While \textsc{GLUE} commonly reports MCC for CoLA, we compute F1 on the acceptability label to keep a
    unified scale across tasks; this benchmark primarily serves as a probe of syntactic/grammatical
    sensitivity.

    \item \textbf{MNLI.}
    Multi-Genre Natural Language Inference is a three-way NLI classification task (entailment/neutral/contradiction)
    spanning diverse text domains.
    It is a stronger, broader probe of inference capability than RTE and is commonly used as a barometer of
    general sentence-pair understanding.

    \item \textbf{MMLU.}
    Measuring Massive Multitask Language Understanding is a multiple-choice benchmark covering a broad set
    of academic and professional subjects.
    Each example presents a question with a fixed set of answer options, testing both factual knowledge and
    multi-step reasoning in a standardized format. 
\end{itemize}

\section{Proofs and Derivations}

\subsection{Definitions and Basic Identities}
\label{app:defs}

Our analysis studies the dynamical evolution of $\| v_n^{\mathrm{old}}\|^2$,
$\| v_n^{\mathrm{new}}\|^2$, and $\| W_n\|_F^2$ under sequential edits.
We first recall the definition of the Frobenius norm and inner product, together with several identities
that will be used repeatedly.

\begin{definition}[Frobenius norm]
\label{def:frob_norm}
For any real matrix $A\in\mathbb{R}^{m\times n}$ with entries $(A)_{ij}=a_{ij}$, the Frobenius norm is
\begin{equation}
\label{eq:frob_norm}
\| A\|_F^2 \;\coloneqq\; \sum_{i=1}^{m}\sum_{j=1}^{n} a_{ij}^2.
\end{equation}
Equivalently, $\| A\|_F$ is the Euclidean norm of $\mathrm{vec}(A)\in\mathbb{R}^{mn}$.
For convenience, we use $\| A\|^2$ to denote $\| A\|_F^2$ when the meaning is clear.
\end{definition}

\begin{definition}[Frobenius inner product]
\label{def:frob_ip}
For any $A,B\in\mathbb{R}^{m\times n}$ with entries $(A)_{ij}=a_{ij}$ and $(B)_{ij}=b_{ij}$, the Frobenius inner product is
\begin{equation}
\label{eq:frob_ip}
\langle A,B\rangle_F \;\coloneqq\; \sum_{i=1}^{m}\sum_{j=1}^{n} a_{ij}\,b_{ij}.
\end{equation}
This coincides with the Euclidean inner product between $\mathrm{vec}(A)$ and $\mathrm{vec}(B)$, and in particular
\begin{equation}
\label{eq:frob_self}
\| A\|_F^2 \;=\; \langle A,A\rangle_F.
\end{equation}
\end{definition}

\begin{lemma}[Useful identities]
\label{lem:frob_identities}
For any $A,B\in\mathbb{R}^{m\times n}$, $u\in\mathbb{R}^{m}$, and $v\in\mathbb{R}^{n}$, the following hold:
\begin{align}
\| A+B\|_F^2
&= \| A\|_F^2 + \| B\|_F^2 + 2\langle A,B\rangle_F, \label{eq:ab_expand}\\
\| u v^\top\|_F^2
&= \| u\|^2\,\| v\|^2, \label{eq:uv_frob}\\
\langle A, u v^\top\rangle_F
&= u^\top A v. \label{eq:auv_ip}
\end{align}
\end{lemma}

\subsection{Proof for Lemma~\ref{lemma3.1}}
\label{app:proof3.1}

Recall that under the convention $C=I$, Eq.~\eqref{eq:deltaW_n} can be written as Eq.~\eqref{eq:deltaW_n_CI}:
\begin{equation}
\Delta W_n
\;=
\frac{\bigl(v_n^{\mathrm{new}}-v_n^{\mathrm{old}}\bigr){k_n^\star}^\top}{\| k_n^\star\|^2}.
\end{equation}

Starting from Eq.~\eqref{eq:W_n_and_W_n-1} and applying the Frobenius identity
\eqref{eq:ab_expand} (Lemma~\ref{lem:frob_identities}), we have
\begin{align}
\|W_n\|^2
&= \| W_{n-1}+\Delta W_n\|^2 \notag\\
&= \|W_{n-1}\|^2 + \| \Delta W_n\|^2 + 2\langle W_{n-1},\Delta W_n\rangle_F .
\label{eq:wnorm_expand}
\end{align}

We now expand the last two terms in Eq.~\eqref{eq:wnorm_expand}.
First, by Eq.~\eqref{eq:uv_frob} (Lemma~\ref{lem:frob_identities}) and \eqref{eq:deltaW_n_CI},
\begin{align}
\| \Delta W_n\|^2
&=
\left\|
\frac{\bigl(v_n^{\mathrm{new}}-v_n^{\mathrm{old}}\bigr){k_n^\star}^\top}{\| k_n^\star\|^2}
\right\|^2 \notag\\
&=
\frac{\| v_n^{\mathrm{new}}-v_n^{\mathrm{old}}\|^2}{\| k_n^\star\|^2}.
\label{eq:deltaW_norm}
\end{align}

Second, by Eq.~\eqref{eq:auv_ip} (Lemma~\ref{lem:frob_identities}) and \eqref{eq:deltaW_n_CI},
\begin{align}
\langle W_{n-1},\Delta W_n\rangle_F
&=
\left\langle
W_{n-1},
\frac{\bigl(v_n^{\mathrm{new}}-v_n^{\mathrm{old}}\bigr){k_n^\star}^\top}{\| k_n^\star\|^2}
\right\rangle_F \notag\\
&=
\frac{1}{\| k_n^\star\|^2}
\left\langle
W_{n-1},
\bigl(v_n^{\mathrm{new}}-v_n^{\mathrm{old}}\bigr){k_n^\star}^\top
\right\rangle_F \notag\\
&=
\frac{1}{\| k_n^\star\|^2}
\bigl(v_n^{\mathrm{new}}-v_n^{\mathrm{old}}\bigr)^\top
W_{n-1}k_n^\star .
\label{eq:deltaW_inner}
\end{align}

Substituting Eqs.~\eqref{eq:deltaW_norm} and~\eqref{eq:deltaW_inner} into
Eq.~\eqref{eq:wnorm_expand} yields
\begin{equation}
\label{eq:wnorm_expand_sub}
\|W_n\|^2
=
\|W_{n-1}\|^2
+
\frac{\| v_n^{\mathrm{new}}-v_n^{\mathrm{old}}\|^2}{\| k_n^\star\|^2}
+
\frac{2}{\| k_n^\star\|^2}
\bigl(v_n^{\mathrm{new}}-v_n^{\mathrm{old}}\bigr)^\top W_{n-1}k_n^\star .
\end{equation}

Recall that $v_n^{\mathrm{old}}=W_{n-1}k_n^\star$, and define $\Delta_n\coloneqq v_n^{\mathrm{new}}-v_n^{\mathrm{old}}$.
Then Eq.~\eqref{eq:wnorm_expand_sub} becomes
\begin{equation}
\label{eq:wnorm_expand_delta}
\|W_n\|^2
=
\|W_{n-1}\|
+
\frac{1}{\| k_n^\star\|^2}
\Bigl(
\| \Delta_n\|^2
+
2\langle \Delta_n, v_n^{\mathrm{old}}\rangle
\Bigr).
\end{equation}

Finally, since $v_n^{\mathrm{new}} = v_n^{\mathrm{old}}+\Delta_n$,
\begin{equation}
\label{eq:delta_identity}
\| \Delta_n\|^2 + 2\langle \Delta_n, v_n^{\mathrm{old}}\rangle
=
\| v_n^{\mathrm{new}}\|^2 - \| v_n^{\mathrm{old}}\|^2.
\end{equation}
Plugging Eq.~\eqref{eq:delta_identity} into Eq.~\eqref{eq:wnorm_expand_delta} gives exactly Eq.~\eqref{eq:lemma3.1},
completing the proof.

\begin{figure*}[t]
  \centering
  \includegraphics[width=0.82\textwidth,clip]{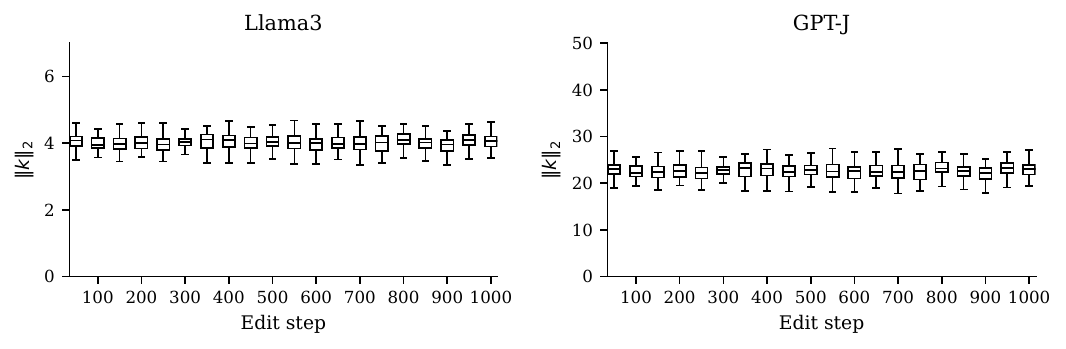}
  \caption{\textbf{$\|k_n^\star\|^2$ stability across edit steps.}
  Boxplots of $\|k_n^\star\|$ over edit steps show small fluctuation and tight concentration,
  supporting the approximation $\|k_n^\star\|^{-2}\approx K$ used in the analysis.}
  \label{fig:stable_k}
\end{figure*}

\begin{figure*}[t]
  \centering
  \includegraphics[width=0.82\textwidth,trim=0 0 0 0,clip]{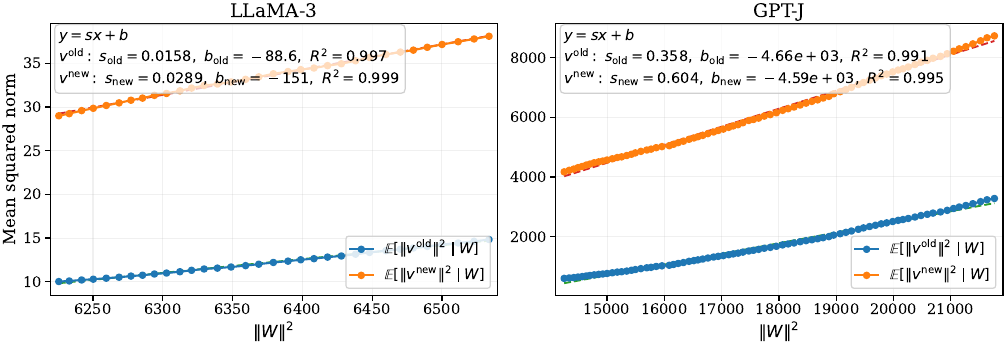}
  \caption{\textbf{Linear scaling between value-vector norms and the edited weight norm (baseline).}
  For sequential editing, we plot $\mathbb{E}\!\left[\| v^{\mathrm{new}}\|^2 \,\middle|\, W\right]$ and $\mathbb{E}\!\left[\| v^{\mathrm{old}}\|^2 \,\middle|\, W\right]$ versus $\| W\|^2$ at multiple checkpoints (averaged over sampled edits/keys). Dashed lines are linear fits; insets report fitted slopes and $R^2$. The fitted slope for $v^{\mathrm{new}}$ is consistently larger than that for $v^{\mathrm{old}}$.}
  \label{fig:Ev-W}
\end{figure*}

\subsection{Proof for Prop.~\ref{prop3.2}}
\label{app:proof3.2}
Starting from Lemma~\ref{lemma3.1} (Eq.~\eqref{eq:lemma3.1}), we take conditional expectation given the current weight $W_{n-1}$:
\begin{equation}
\label{eq:prop3.2_condexp_start}
\mathbb{E}\!\left[\|W_n\|^2 \,\middle|\, W_{n-1}\right]
=
\|W_{n-1}\|^2
+
\mathbb{E}\!\left[
\frac{\|v_n^{\mathrm{new}}\|^2-\|v_n^{\mathrm{old}}\|^2}{\|k_n^\star\|^2}
\,\middle|\,
W_{n-1}
\right].
\end{equation}

Empirically, $\|k_n^\star\|$ varies little across edit steps (Fig.~\ref{fig:stable_k}).
We therefore treat $\|k_n^\star\|^{-2}$ as approximately constant and set
\begin{equation}
K \;:=\; \text{the empirical mean of }\ \|k_n^\star\|^{-2}\ \text{over the sampled checkpoints and sampled edits/keys}.
\label{eq:k_n_approx}
\end{equation}

Substituting $\|k_n^\star\|^{-2}\approx K$ into \eqref{eq:prop3.2_condexp_start} yields

\begin{equation}
\mathbb{E}\!\left[\|W_n\|^2 \,\middle|\, W_{n-1}\right]
\;\approx\;
\|W_{n-1}\|^2 + K\cdot
\mathbb{E}\!\left[\|v_n^{\mathrm{new}}\|^2-\|v_n^{\mathrm{old}}\|^2 \,\middle|\, W_{n-1}\right].
\label{eq:prop3.2_K-approxed}
\end{equation}

Next, Fig.~\ref{fig:Ev-W} shows an approximately linear scaling between value-vector norms and the edited weight norm over the observation range:
\begin{align}
\label{eq:linear_fit_new}
\mathbb{E}\!\left[\|v_n^{\mathrm{new}}\|^2 \,|\, W_{n-1}\right]
&\approx
s_{\mathrm{new}}\,\|W_{n-1}\|^2 + b_{\mathrm{new}},\\
\label{eq:linear_fit_old}
\mathbb{E}\!\left[\|v_n^{\mathrm{old}}\|^2 \,\middle|\, W_{n-1}\right]
&\approx
s_{\mathrm{old}}\,\|W_{n-1}\|^2 + b_{\mathrm{old}},
\end{align}
where $s_{\mathrm{new}} > s_{\mathrm{old}} > 0$ .

Subtracting Eq.~\eqref{eq:linear_fit_old} from Eq.~\eqref{eq:linear_fit_new} gives
\begin{equation}
\label{eq:diff_linear}
\mathbb{E}\!\left[
\|v_n^{\mathrm{new}}\|^2-\|v_n^{\mathrm{old}}\|^2
\,\middle|\,
W_{n-1}
\right]
\approx
(s_{\mathrm{new}}-s_{\mathrm{old}})\,\|W_{n-1}\|^2 + (b_{\mathrm{new}}-b_{\mathrm{old}}).
\end{equation}

For notational convenience, we introduce a unified reparametrization of the empirical linear fits by the two constants $(\rho,\gamma)$:
\begin{equation}
\label{eq:rho_gamma_map}
\rho(s_{\mathrm{new}},s_{\mathrm{old}};K)
:= 1 + K(s_{\mathrm{new}}-s_{\mathrm{old}}) ,
\qquad
\gamma(s_{\mathrm{new}},b_{\mathrm{new}},s_{\mathrm{old}},b_{\mathrm{old}})
:= \frac{b_{\mathrm{new}}-b_{\mathrm{old}}}{s_{\mathrm{new}}-s_{\mathrm{old}}} .
\end{equation}

Then Eq.~\eqref{eq:diff_linear} can be rewritten as
\begin{equation}
\label{eq:diff_linear_rho_gamma}
\mathbb{E}[\|v_n^{\mathrm{new}}\|^2-\|v_n^{\mathrm{old}}\|^2 \mid W_{n-1}]
\approx
(s_{\mathrm{new}}-s_{\mathrm{old}})\,(\|W_{n-1}\|^2 + \gamma) ,\quad where\, \gamma=\gamma(s_{\mathrm{new}},b_{\mathrm{new}},s_{\mathrm{old}},b_{\mathrm{old}})
\end{equation}

Substituting Eq.~\eqref{eq:diff_linear_rho_gamma} into Eq.~\eqref{eq:prop3.2_K-approxed} yields
\begin{equation}
\label{eq:prop3.2_condexp_recursion}
\mathbb{E}[\|W_n\|^2 \mid W_{n-1}]
\approx
\|W_{n-1}\|^2 + K(s_{\mathrm{new}}-s_{\mathrm{old}})\,(\|W_{n-1}\|^2 + \gamma)
=
\rho\,\|W_{n-1}\|^2 + (\rho-1)\gamma , \quad where \,\rho=\rho(s_{\mathrm{new}},s_{\mathrm{old}};K).
\end{equation}

Taking expectation on both sides and using the tower property,
\begin{equation}
\label{eq:prop3.2_uncond_recursion}
\mathbb{E}\|W_n\|^2
\approx
\rho\,\mathbb{E}\|W_{n-1}\|^2 + (\rho-1)\gamma .
\end{equation}

Since $K>0$ and $s_{\mathrm{new}}-s_{\mathrm{old}}>0$, we have $\rho>1$.
Solving Eq.~\eqref{eq:prop3.2_uncond_recursion} gives
\begin{equation}
\label{eq:prop3.2_closed_form_rho_gamma}
\mathbb{E}\|W_n\|^2
\approx
\rho^n\,\mathbb{E}\|W_0\|^2 + \gamma(\rho^n-1) .
\end{equation}

In the main text, we denote the baseline (divergent) regime by $(\alpha,R)$, which is the specialization of $(\gamma,\rho)$ under the empirical fits:
\begin{equation}
\label{eq:align_alpha_R}
R := \rho(s_{\mathrm{new}},s_{\mathrm{old}};K) ,
\qquad
\alpha := \gamma(s_{\mathrm{new}},b_{\mathrm{new}},s_{\mathrm{old}},b_{\mathrm{old}}) .
\end{equation}

Using notations above, Eq.~\eqref{eq:prop3.2_closed_form_rho_gamma} can be rewritten as:
\begin{equation}
\label{eq:prop3.2_closed_form_alpha_R}
\mathbb{E}\|W_n\|^2
\approx
R^n\,\mathbb{E}\|W_0\|^2 + \alpha(R^n-1) ,
\end{equation}
which is exactly Eq.~(12). In particular, $ R>1$ indicates an approximately exponential growth trend in $\mathbb{E}\|W_n\|^2$ under the baseline empirical fits.
 We will introduce an alternative parametrization $(\beta,r)$ for the NAS (stable) regime in the next subsection.

\subsection{Proof for Cor.~\ref{cor3.3}} 
\label{app:proof3.3}
We now analyze our method, which introduces a negative feedback by enforcing a constant target norm for the post-edit value vector.
Specifically, we impose
\begin{equation}
\label{eq:vnew_NAS}
\|v_n^{\mathrm{new}}\|^2 := \tau^2,\qquad \tau>0 .
\end{equation}
Equivalently,
\begin{equation}
\label{eq:vnew_NAS_condexp}
\mathbb{E}[\|v_n^{\mathrm{new}}\|^2 \mid W_{n-1}] = \tau^2 .
\end{equation}
Using the same linear-fit notation as in Appendix~\ref{app:proof3.2}, this corresponds to the specialization
\begin{equation}
\label{eq:NAS_specialization}
s_{\mathrm{new}}=0, \qquad b_{\mathrm{new}}=\tau^2 .
\end{equation}

\begin{figure*}[t]
  \centering
  \includegraphics[width=0.82\textwidth]{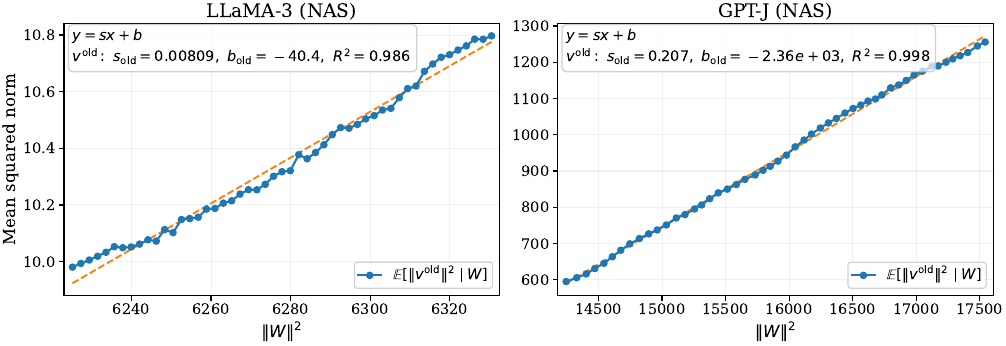}
  \vspace{-0.5em}
  \caption{\textbf{Linear scaling between the pre-edit value norm and the edited weight norm (NAS).}
    Under norm anchoring, we plot $\mathbb{E}[\|v_n^{\mathrm{old}}\|^2 \mid W]$ versus $\|W\|^2$ at multiple checkpoints (averaged over sampled edits/keys). Dashed lines are linear fits; insets report fitted slopes and $R^2$.}
  \label{fig:Ev-W_NAS}
  \vspace{-0.5em}
\end{figure*}

Meanwhile, within the same observation range (Fig.~\ref{fig:Ev-W_NAS}), we empirically find an approximately stable linear scaling for the pre-edit value norm:
\begin{equation}
\label{eq:linear_fit_old_NAS}
\mathbb{E}\|v_n^{\mathrm{old}}\|^2
\approx
s_{\mathrm{old}}\,\mathbb{E}\|W_{n-1}\|^2 + b_{\mathrm{old}} .
\end{equation}

Recall the unified parametrization in Appendix~\ref{app:proof3.2} (Eq.~\eqref{eq:rho_gamma_map}).
Under the NAS specialization $s_{\mathrm{new}}=0$ and $b_{\mathrm{new}}=\tau^2$, we have
\begin{equation}
\label{eq:rho_gamma_NAS}
\rho = \rho(0,s_{\mathrm{old}};K) = 1 - Ks_{\mathrm{old}} ,
\qquad
\gamma = \gamma(0,\tau^2,s_{\mathrm{old}},b_{\mathrm{old}})
= \frac{\tau^2-b_{\mathrm{old}}}{0-s_{\mathrm{old}}} .
\end{equation}

Following the same derivation as in Appendix~B.3, substituting $\|v_n^{\mathrm{new}}\|^2 \mapsto \tau^2$ and using Eq.~\eqref{eq:linear_fit_old_NAS} yields
\begin{equation}
\label{eq:cor3.3_uncond_recursion_rho_gamma}
\mathbb{E}\|W_n\|^2
\approx
\rho\,\mathbb{E}\|W_{n-1}\|^2 + (\rho-1)\gamma .
\end{equation}

Solving Eq.~\eqref{eq:cor3.3_uncond_recursion_rho_gamma} gives
\begin{equation}
\label{eq:cor3.3_closed_form_rho_gamma}
\mathbb{E}\|W_n\|^2
\approx
\rho^n\,\mathbb{E}\|W_0\|^2 + \gamma(\rho^n-1) .
\end{equation}

In the main text, we reparametrize the NAS (stable) regime by $(\beta,r)$ to highlight boundedness:
\begin{equation}
\label{eq:align_beta_r}
r := \rho(0,s_{\mathrm{old}};K)=1-Ks_{\mathrm{old}} ,
\qquad
\beta := -\gamma(0,\tau^2,s_{\mathrm{old}},b_{\mathrm{old}}) = \frac{\tau^2-b_{\mathrm{old}}}{s_{\mathrm{old}}} .
\end{equation}

Substituting Eq.~\eqref{eq:align_beta_r} into Eq.~\eqref{eq:cor3.3_closed_form_rho_gamma} yields
\begin{equation}
\label{eq:cor3.3_closed_form_beta_r}
\mathbb{E}\|W_n\|^2
\approx
r^n\,\mathbb{E}\|W_0\|^2 + \beta(1-r^n) .
\end{equation}

which is exactly Eq.~\eqref{eq:cor3.3}. In our experiments, the fitted slope satisfies $s_{\mathrm{old}}>0$ and the estimated product $Ks_{\mathrm{old}}$ lies in $(0,1)$ over the observation range, hence $0<r=1-Ks_{\mathrm{old}}<1$.
Moreover, we typically observe $b_{\mathrm{old}}<0$ and set $\tau>0$, so $\beta=\frac{\tau^2-b_{\mathrm{old}}}{s_{\mathrm{old}}}>0$.
Therefore, Eq.~\eqref{eq:cor3.3_closed_form_beta_r} suggests that $\mathbb{E}\|W_n\|^2$ remains uniformly bounded over $n$.

\subsection[General Case for C not equal I]{General Case for $C\neq I$}
\label{app:proofgeneral}

\paragraph{\textbf{Lemma 3.1 (General $C$ Case).}}
For the general $C$ case, Eq.~\eqref{eq:lemma3.1} in Lemma~\ref{lemma3.1} becomes
\begin{equation}
\label{eq:lemma3.1_tilde}
\boxed{
\|\tilde W_n\|^2
=
\|\tilde W_{n-1}\|^2
+\frac{\|v_n^{\mathrm{new}}\|^2-\|v_n^{\mathrm{old}}\|^2}{\|\tilde k_n\|^2}
},
\end{equation}
where
\begin{equation}
\label{eq:tilde_defs}
\tilde W_n \;\coloneqq\; W_n C^{1/2},
\qquad
\Delta \tilde W_n \;\coloneqq\; \Delta W_n C^{1/2},
\qquad
\tilde k_n \;\coloneqq\; C^{-1/2}k_n^\star.
\end{equation}

\paragraph{Proof:}
Consider Eq.~\eqref{eq:deltaW_n}:
\begin{equation}
\label{eq:deltaW_generalC}
\Delta W_n=\frac{(v_n^{\mathrm{new}}-v_n^{\mathrm{old}})(C^{-1}k_n^\star)^{\top}}{{k_n^\star}^{\top}C^{-1}k_n^\star}.
\end{equation}

For the denominator, note that
\begin{equation}
{k_n^\star}^{\top}C^{-1}k_n^\star
=
(C^{-1/2}k_n^\star)^{\top}(C^{-1/2}k_n^\star)
=
\tilde k_n^{\top}\tilde k_n
=
\|\tilde k_n\|^2.
\end{equation}

For the numerator, observe that
\begin{equation}
(C^{-1}k_n^\star)^{\top}
=
\bigl(C^{-1/2}(C^{-1/2}k_n^\star)\bigr)^{\top}
=
(C^{-1/2}\tilde k_n)^{\top}
=
\tilde k_n^{\top}(C^{-1/2})^{\top}
=
\tilde k_n^{\top}C^{-1/2}.
\end{equation}
Therefore,
\begin{equation}
\label{eq:deltaW_generalC_rewrite}
\Delta W_n
=
\frac{\left(v_n^{\mathrm{new}}-v_n^{\mathrm{old}}\right)\tilde{k}_n^{\top}C^{-1/2}}{\left\|\tilde{k}_n\right\|^2}.
\end{equation}

Since $W_n=W_{n-1}+\Delta W_n$, multiplying both sides by $C^{1/2}$ gives
\begin{equation}
\label{eq:W_n_and_W_n-1_tilde}
\tilde W_n = \tilde W_{n-1} + \Delta \tilde W_n,
\end{equation}
where
\begin{equation}
\label{eq:deltaW_tilde}
\Delta \tilde W_n
=\Delta W_n C^{1/2}
=
\frac{\left(v_n^{\mathrm{new}}-v_n^{\mathrm{old}}\right)\tilde{k}_n^{\top}}{\left\|\tilde{k}_n\right\|^2}.
\end{equation}

Eqs.~\eqref{eq:W_n_and_W_n-1_tilde} and~\eqref{eq:deltaW_tilde} have the same form as Eqs.~\eqref{eq:W_n_and_W_n-1} and~\eqref{eq:deltaW_n_CI} under $C=I$.
Thus, by replacing $(W_n,k_n^\star)$ with $(\tilde W_n,\tilde k_n)$ in Appendix~\ref{app:proof3.1}, we obtain

\begin{equation}
\|\tilde W_n\|^2
=
\|\tilde W_{n-1}\|^2
+\frac{\|v_n^{\mathrm{new}}\|^2-\|v_n^{\mathrm{old}}\|^2}{\|\tilde k_n\|^2},
\end{equation}

which completes the proof.

\paragraph{\textbf{Propositon 3.2 (General $C$ Case).}}
For general $C$ case, Eq.~\eqref{prop3.2} in Prop.~\ref{prop3.2} becomes
\begin{equation}
\label{eq:prop3.2_tilde}
\boxed{
\mathbb{E}\|\tilde W_n\|^2
\approx
R^n\,\mathbb{E}\|\tilde W_0\|^2 + \alpha(R^n-1) 
},
\end{equation}

where $R>1$.
The derivation is identical to Appendix~\ref{app:proof3.2}, by replacing $W_n,k_n^\star$ with the ``tilde'' versions defined in Eq.~\eqref{eq:tilde_defs}.

\paragraph{Proof:}
Recall that the proof of Prop.\ref{prop3.2} directly begins with Eq.~\eqref{eq:lemma3.1}, which has the same form as Eq.~\eqref{eq:lemma3.1_tilde} above. Therefore we can very well reuse the derivation in Prop.\ref{prop3.2} line-by-line. The only caveat here is whether the empirical observations in Fig.~\ref{fig:stable_k}, Fig.~\ref{fig:Ev-W},  still hold under the ``tilde'' transformation.
Concretely, the two approximations used in Appendix~\ref{app:proof3.2} should be replaced by:

\begin{figure*}[t]
  \centering
  \includegraphics[width=0.82\textwidth,clip]{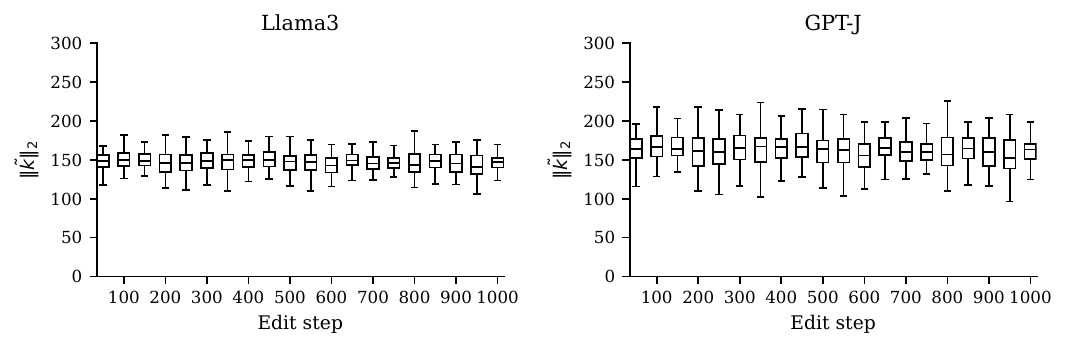}
  \vspace{-0.3em}
  \caption{\textbf{$\|\tilde k_n\|^2$ stability across edit steps.}
  Boxplots of $\|\tilde k_n\|$ over edit steps show small fluctuation and tight concentration,
  supporting the approximation $\|\tilde k_n\|^{-2}\approx \tilde K$ used in the analysis.}
  \label{fig:stable_k_tilde}
  \vspace{-0.3em}
\end{figure*}

\begin{figure*}[t]
  \centering
  \includegraphics[width=0.82\textwidth,trim=0 0 0 0,clip]{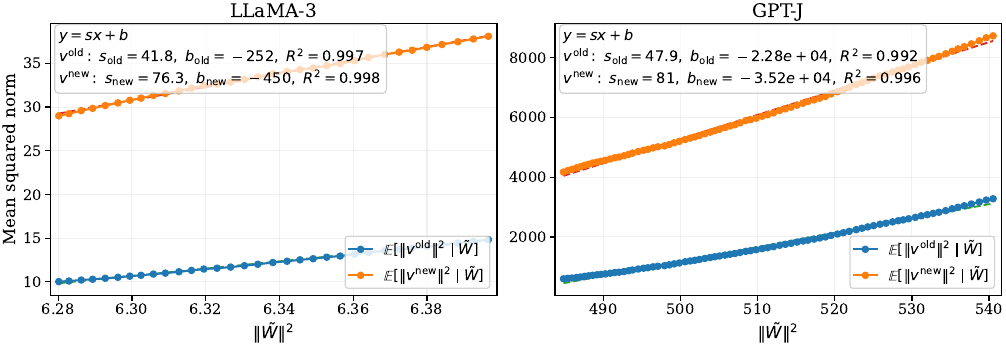}
  \vspace{-0.5em}
  \caption{\textbf{Linear scaling between value-vector norms and the edited weight norm (baseline, tilde space).}
  For sequential editing, we plot $\mathbb{E}\!\left[\| v^{\mathrm{new}}\|^2 \,\middle|\, \tilde W\right]$ and $\mathbb{E}\!\left[\| v^{\mathrm{old}}\|^2 \,\middle|\, \tilde W\right]$ versus $\| \tilde W\|^2$ at multiple checkpoints (averaged over sampled edits/keys). Dashed lines are linear fits; insets report fitted slopes and $R^2$. The fitted slope for $v^{\mathrm{new}}$ is consistently larger than that for $v^{\mathrm{old}}$.}
  \vspace{-0.3em}
  \label{fig:Ev-W_tilde}
\end{figure*}

\begin{enumerate}
\item \textbf{$\|\tilde k_n\|^2$ Stability (Fig.~\ref{fig:stable_k_tilde}).}
Replace
\[
\|k_n^\star\|^{-2}\approx K
\]
by
\[
\|\tilde k_n\|^{-2}\approx \tilde K,
\]
i.e., $\|\tilde k_n\|^{-2}$ remains stable around a constant during sequential edits.

\item \textbf{Linear fits (Fig.~\ref{fig:Ev-W_tilde}).}
Replace
\[
\mathbb{E}\!\left[\|v_n^{\mathrm{new}}\|^2 \,\middle|\, \|W_{n-1}\|^2\right]\approx s_{\mathrm{new}}\|W_{n-1}\|^2+b_{\mathrm{new}},
\qquad
\mathbb{E}\!\left[\|v_n^{\mathrm{old}}\|^2 \,\middle|\, \|W_{n-1}\|^2\right]\approx s_{\mathrm{old}}\|W_{n-1}\|^2+b_{\mathrm{old}}
\]
by the corresponding relations for the ``tilde'' versions:
\[
\mathbb{E}\!\left[\|v_n^{\mathrm{new}}\|^2 \,\middle|\, \|\tilde W_{n-1}\|^2\right]\approx s_{\mathrm{new}}\|\tilde W_{n-1}\|^2+b_{\mathrm{new}},
\qquad
\mathbb{E}\!\left[\|v_n^{\mathrm{old}}\|^2 \,\middle|\, \|\tilde W_{n-1}\|^2\right]\approx s_{\mathrm{old}}\|\tilde W_{n-1}\|^2+b_{\mathrm{old}}.
\]
\end{enumerate}
Under these empirical observations for the ``tilde'' versions, we can define $(\alpha,R)$ (resp. $(\beta,r)$) for the general $C$ case. And the derivation in Appendix~\ref{app:proof3.2} can be reused line-by-line, yielding Eq.~\eqref{eq:prop3.2_tilde}.

\begin{figure}[H]
  \centering
  \vspace{-0.5mm}
  \includegraphics[width=0.82\textwidth]{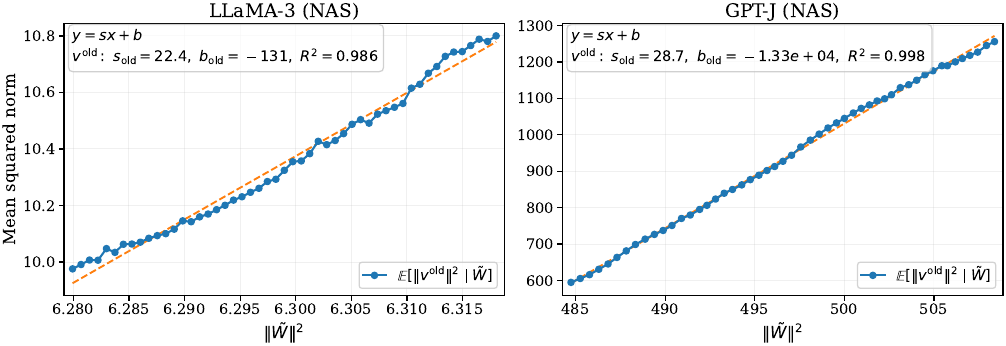}
  \vspace{-0.3em}
  \caption{\textbf{Linear scaling between the pre-edit value norm and the edited weight norm (NAS, tilde space).}
  Under norm anchoring, we plot $\mathbb{E}[\|v_n^{\mathrm{old}}\|^2 \mid \tilde W]$ versus $\|\tilde W\|^2$ at multiple checkpoints (averaged over sampled edits/keys). Dashed lines are linear fits; insets report fitted slopes and $R^2$.}
  \label{fig:Ev-W_NAS_tilde}
  \vspace{-0.3em}
\end{figure}

\paragraph{\textbf{Corollary 3.3 (General $C$ Case).}}
For general $C$ case, Eq.~\eqref{cor3.3} in Prop.~\ref{cor3.3} becomes
\begin{equation}
\label{eq:cor3.3_tilde}
\boxed{
\mathbb{E}\|\tilde W_n\|^2
\approx
r^n\,\mathbb{E}\|\tilde W_0\|^2 + \beta(1-r^n) 
},
\end{equation}
where $\beta >0,\quad r \in (0,1)$.
\paragraph{Proof:}
Similar to \textbf{Proposition 3.2 (General $C$ Case)}, to reuse the derivation in Appendix~\ref{app:proof3.3}, it suffices to verify that the after-NAS empirical fit used in Appendix~\ref{app:proof3.3} remains valid under the transformation in Eq.~\eqref{eq:tilde_defs}.
Specifically, the approximately stable relation in the observation range,
\[
\mathbb{E}\|v_n^{\mathrm{old}}\|^2 \approx s_{\mathrm{old}}\,\mathbb{E}\|W_{n-1}\|^2 + b_{\mathrm{old}},
\]
should be replaced in the general $C$ case by: (Fig.~\ref{fig:Ev-W_NAS_tilde})

\[
\mathbb{E}\|v_n^{\mathrm{old}}\|^2 \approx s_{\mathrm{old}}\,\mathbb{E}\|\tilde W_{n-1}\|^2 + b_{\mathrm{old}}.
\]
Since the above ``tilde'' empirical phenomenon also holds, the remaining computations in Appendix~\ref{app:proof3.3} can be reused line-by-line, yielding the corresponding conclusion of Cor.~\ref{cor3.3} for the general $C$ case.

\section{Additional Experiment}
\label{app:c}

\subsection{Single-Edit Plasticity under Norm Anchoring}
\label{app:single_edit}

RQ4 examines whether the long-horizon gains of NAS are obtained by weakening individual edits.
To isolate this question from sequential accumulation effects, we evaluate independent single edits:
each edit starts from the original base model, applies one edit, and is then evaluated in isolation.
We use 1,000 CounterFact examples on Llama-3-8B and compare each base editor with and without
NAS under the same editing configuration.

Table~\ref{tab:app_single_edit_all} reports the full single-edit results.
NAS does not systematically reduce single-edit efficacy. For MEMIT, the overall score increases
from 88.46 to 89.16; for AlphaEdit, it increases from 88.03 to 89.36.
The locality metric changes only marginally, suggesting that anchoring the final solved value vector
does not introduce a broad degradation of edit locality in the single-edit setting.

\begin{table}[t]
\centering
\small
\caption{
Independent single-edit results on 1,000 CounterFact examples with Llama-3-8B.
Each edit starts from the original base model and is evaluated in isolation.
NAS does not systematically reduce single-edit efficacy.
}
\label{tab:app_single_edit_all}
\begin{tabular}{lccccc}
\toprule
Editor & NAS & Rewrite Succ. & Paraphrase Succ. & Locality Succ. & Score \\
\midrule
MEMIT     & Off & 98.20 & 80.30 & 88.68 & 88.46 \\
MEMIT     & On  & 98.20 & 82.10 & 88.62 & 89.16 \\
AlphaEdit & Off & 98.00 & 81.00 & 86.74 & 88.03 \\
AlphaEdit & On  & 99.00 & 83.90 & 86.56 & 89.36 \\
\bottomrule
\end{tabular}
\end{table}

We further test a harder subset where NAS applies the strongest magnitude correction.
Specifically, for each base editor, we select the top 25\% examples whose solved value vectors
have the largest norms relative to the NAS anchor, i.e., the cases that receive the strongest
downward rescaling under NAS.
This subset directly probes whether norm anchoring sacrifices edit plasticity when the intervention
is most active.

Table~\ref{tab:app_single_edit_top25} shows that the effect remains limited even in this hard subset.
For AlphaEdit, NAS is neutral to slightly beneficial, increasing the score from 91.03 to 91.27.
For MEMIT, NAS incurs a modest score decrease from 92.98 to 91.81, mainly through paraphrase
success, while rewrite success remains very high and locality slightly improves.
Thus, any single-edit plasticity cost is small and localized, and cannot explain the large
long-horizon gains observed in the sequential setting.

\begin{center}
\begin{minipage}{\textwidth}
\centering
\small
\captionof{table}{
Single-edit results on the top 25\% most strongly downscaled CounterFact examples.
These are the cases whose solved value vectors have the largest norms relative to the NAS anchor.
Even in this hard subset, NAS does not show a broad collapse of single-edit efficacy.
}
\label{tab:app_single_edit_top25}
\begin{tabular}{lccccc}
\toprule
Editor & NAS & Eff. & Gen. & Loc. & Score \\
\midrule
MEMIT     & Off & 100.00 & 90.80 & 88.88 & 92.98 \\
MEMIT     & On  & 99.20  & 88.00 & 89.04 & 91.81 \\
AlphaEdit & Off & 99.20  & 88.40 & 86.48 & 91.03 \\
AlphaEdit & On  & 99.20  & 88.40 & 87.12 & 91.27 \\
\bottomrule
\end{tabular}
\end{minipage}
\end{center}

Overall, these single-edit analyses support the interpretation in RQ4: NAS's long-horizon
improvements are not primarily driven by a broad stability--plasticity trade-off.
Instead, anchoring the final solved value vector can improve sequential stability while leaving
standard single-edit behavior largely intact.

\subsection{Robustness to Anchor Mis-Specification}
\label{app:anchor_misspec}

NAS uses an original-model reference norm as the anchor. 
A natural question is whether the method depends on a narrowly tuned anchor value. 
To test this, we intentionally bias the anchor estimate by selecting pilot subsets with lower or higher solved-value norms, and then run 4,000 sequential CounterFact~\citep{meng2022rome} edits under the same configuration.

\begin{center}
\begin{minipage}{\textwidth}
\centering
\small
\captionof{table}{
Robustness to anchor mis-specification. 
The target norm is intentionally estimated from lower- or higher-norm pilot subsets.
}
\label{tab:app_anchor_misspec}
\begin{tabular}{l c c}
\toprule
Anchor source & Target norm & Score@4000 \\
\midrule
Bottom 10\% & 4.6434 & 83.79 \\
Bottom 25\% & 4.8226 & 83.88 \\
Default     & 5.3161 & 83.44 \\
Top 25\%    & 5.7927 & 82.98 \\
Top 10\%    & 5.9504 & 83.07 \\
\bottomrule
\end{tabular}
\end{minipage}
\end{center}

The anchor varies from 4.6434 to 5.9504, while the final score changes by only 0.90 points. 
Thus, NAS does not rely on a precisely tuned anchor. 
The main effect is to keep solved value vectors within a bounded operating range, rather than to select one uniquely optimal target norm.

\subsection{Robustness under Non-Stationary Edit Orders}
\label{app:order_shift}

A fixed anchor could be brittle if the edit stream is non-stationary, for example when easy and difficult edits appear in different phases. 
To stress-test this setting, we reorder the first 4,000 CounterFact edits using a difficulty proxy computed once on the original model: the native MEMIT write norm. 
We compare three orders: the default stream order, low-to-high difficulty, and high-to-low difficulty.

\begin{center}
\begin{minipage}{\textwidth}
\centering
\small
\captionof{table}{
Robustness under non-stationary edit orders. 
Scores are reported at a key horizon for each editor, together with the range across three edit orders. 
The later-horizon column reports scores under default / low-to-high / high-to-low orders.
}
\label{tab:app_order_shift}
\resizebox{\textwidth}{!}{%
\begin{tabular}{l c c c c c c l}
\toprule
Editor & NAS & Key horizon & Default & Low$\rightarrow$High & High$\rightarrow$Low & Range & Later horizon \\
\midrule
AlphaEdit & Off & 1500 & 81.76 & 82.77 & 54.70 & 28.07 & @4000: 52.85 / 50.32 / 50.76 \\
AlphaEdit & On  & 1500 & 86.25 & 84.91 & 87.68 & 2.77  & @4000: 83.44 / 82.01 / 83.76 \\
MEMIT     & Off & 300  & 85.87 & 84.88 & 51.52 & 34.35 & @1000: 50.57 / 50.54 / 50.37 \\
MEMIT     & On  & 300  & 90.56 & 90.76 & 90.94 & 0.38  & @1000: 80.32 / 78.98 / 68.59 \\
\bottomrule
\end{tabular}
}
\end{minipage}
\end{center}

NAS substantially reduces order sensitivity. 
For AlphaEdit, the cross-order spread at 1,500 edits decreases from 28.07 to 2.77 after enabling NAS; for MEMIT, the spread at 300 edits decreases from 34.35 to 0.38. 
Under the harshest high-to-low order, MEMIT+NAS still degrades by 1,000 edits, but remains well above the collapsed baseline. 
These results support the robustness of a fixed oridinal-model anchor while also motivating adaptive anchoring as a possible extension under severe non-stationarity.

\subsection{Multi-Hop Counterfactual Editing}
\label{app:multihop}

We also evaluate NAS on MQuAKE~\citep{zhong-etal-2023-mquake}, a counterfactual multi-hop editing benchmark. 
This setting probes whether NAS can stabilize sequential editing without causing an early under-editing penalty on harder reasoning-oriented edits.

\begin{center}
\begin{minipage}{\textwidth}
\centering
\small
\captionof{table}{
Sequential multi-hop editing results on MQuAKE-CF-3k-v2. 
``First $<5$'' denotes the first edit step where accuracy falls below 5\%.
}
\label{tab:app_multihop}
\begin{tabular}{lcccc}
\toprule
Method & Acc@100 & Peak & Acc@3000 & First $<5$ \\
\midrule
NAS       & 16.68 & 19.87@900 & 17.46 & -- \\
AlphaEdit & 16.17 & 19.28@700 & 0.04  & 1400 \\
MEMIT     & 14.49 & 14.49@100 & 0.00  & 200  \\
ROME      & 3.33  & 3.33@100  & 0.00  & 100  \\
\bottomrule
\end{tabular}
\end{minipage}
\end{center}

NAS shows no early under-editing penalty on this multi-hop benchmark: its Acc@100 is comparable to AlphaEdit and higher than MEMIT and ROME. 
More importantly, NAS maintains 17.46\% accuracy at 3,000 edits, whereas AlphaEdit, MEMIT, and ROME collapse to near-zero accuracy. 
This supports the view that norm anchoring improves long-run stability without simply weakening the edit signal.

\subsection{Additional Short-Stream KnowEdit Benchmarks}
\label{app:short_knowedit}

To complement the long-stream benchmarks used in the main text, we also evaluate NAS on two shorter KnowEdit-style benchmarks, WikiRecent and WikiBio~\citep{zhang2024comprehensive}, using Llama-3-8B. 
These datasets are not intended to stress the ultra-long regime, but they provide additional breadth over more recent and biographical factual updates. 
Table~\ref{tab:app_wikirecent_wikibio} reports end-of-stream results.

\begin{center}
\begin{minipage}{\textwidth}
\centering
\small
\captionof{table}{
End-of-stream results on additional KnowEdit benchmarks. 
All numbers are percentages. WikiBio does not report paraphrase success under its protocol.
}
\label{tab:app_wikirecent_wikibio}
\begin{tabular}{l l c c c}
\toprule
Dataset & Method & Rewrite & Paraphrase & Locality \\
\midrule
WikiRecent (1266) & NAS       & 96.65 & 62.76 & 73.20 \\
WikiRecent (1266) & AlphaEdit & 95.07 & 61.80 & 70.77 \\
WikiRecent (1266) & MEMIT     & 0.00  & 0.00  & 0.00  \\
WikiRecent (1266) & PRUNE     & 0.00  & 0.00  & 0.00  \\
WikiRecent (1266) & RECT      & 0.00  & 0.00  & 0.00  \\
\midrule
WikiBio test (306) & NAS       & 71.81 & -- & 35.71 \\
WikiBio test (306) & AlphaEdit & 62.14 & -- & 33.04 \\
WikiBio test (306) & MEMIT     & 0.00  & -- & 0.00  \\
WikiBio test (306) & PRUNE     & 0.00  & -- & 0.00  \\
WikiBio test (306) & RECT      & 0.00  & -- & 0.00  \\
\bottomrule
\end{tabular}
\end{minipage}
\end{center}

NAS remains stable to the end of both short streams and outperforms AlphaEdit on all reported metrics. 
The results suggest that the gains of NAS are not limited to the full CounterFact, ZsRE, or WikiBigEdit streams used in the main experiments.

\subsection{Additional General Capability Experiment}
\label{app:extrageneral}

\subsubsection{Qwen2.5}
\label{generalqwen}

\begin{figure}[H]
    \centering
    \includegraphics[width=0.82\textwidth]{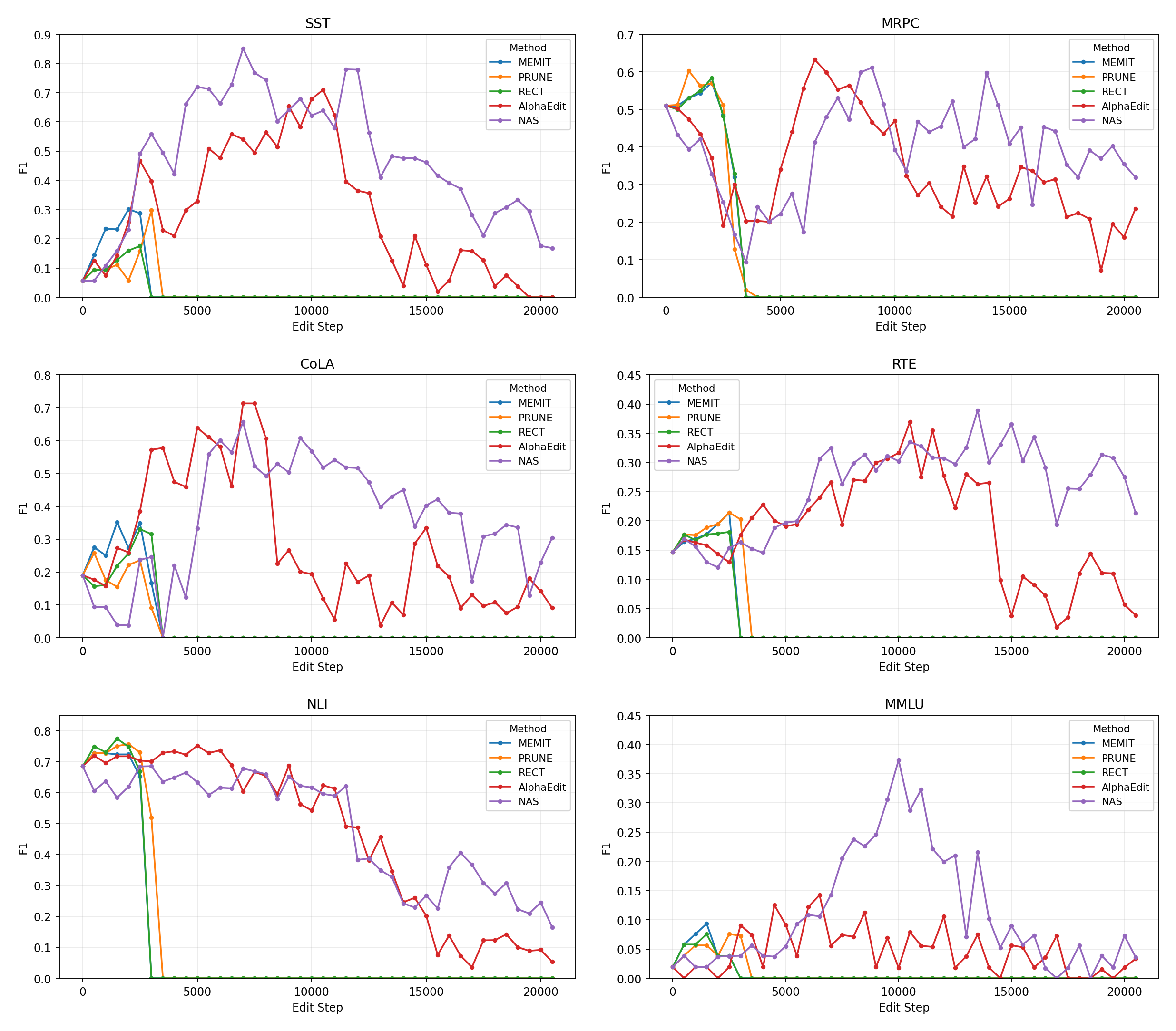}
    \caption{GLUE F1 trajectories over sequential edits on Qwen (step 0 denotes the pre-edit baseline).}
    \label{fig:appendix:glue_qwen_f1_traj}
\end{figure}

Appendix Fig.~\ref{fig:appendix:glue_qwen_f1_traj} reports the evolution of GLUE F1 on Qwen as the number of edits increases (step~0 is the pre-edit baseline).
Across all six tasks, \textsc{NAS} yields consistently stronger retention of general language understanding compared with prior locate-and-edit editors, with the gap
becoming most visible in the long-horizon regime where cumulative interference typically dominates.
This behavior supports our central claim that \textsc{NAS} acts as a plug-in stabilizer for lifelong knowledge updating: by controlling the destabilizing
drift induced by repeated localized rewrites, it substantially reduces collateral degradation on out-of-distribution evaluation suites such as GLUE.

\subsubsection{GPT-J}
\label{generalgptj}

\begin{figure}[H]
    \centering
    \includegraphics[width=0.82\textwidth]{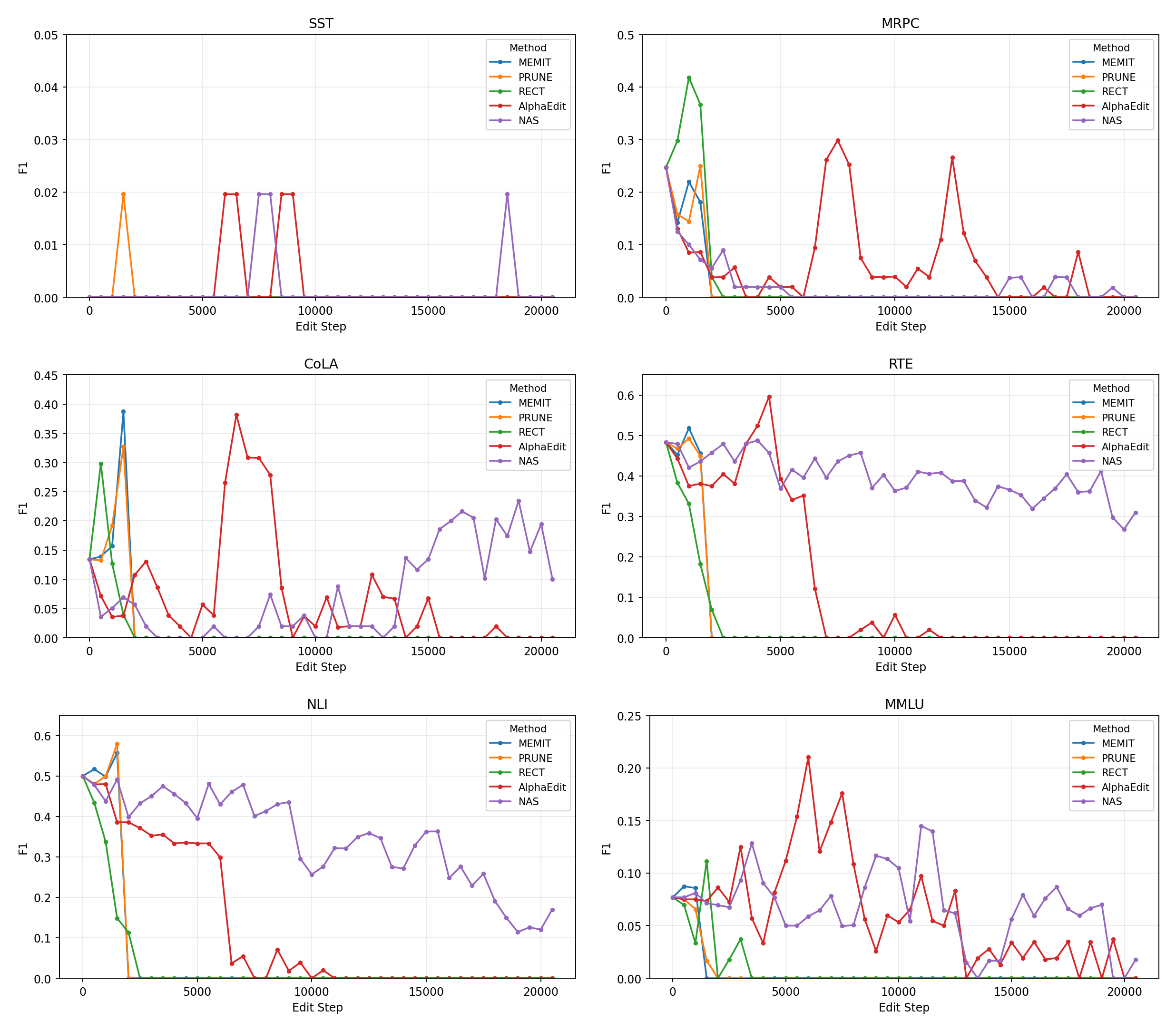}
    \caption{GLUE F1 trajectories over sequential edits on GPT-J (step 0 denotes the pre-edit baseline).}
    \label{fig:appendix:glue_gptj_f1_traj}
\end{figure}

A similar trend is observed on GPT-J in Appendix Fig.~\ref{fig:appendix:glue_gptj_f1_traj}, where the backbone is generally more fragile under repeated edits.
Despite the lower absolute starting performance on some tasks, \textsc{NAS} markedly improves long-horizon stability, maintaining non-trivial GLUE performance
in stages where competing editors often exhibit sharp degradation.
The fact that the same plug-in provides clear benefits on both Qwen and GPT-J indicates that \textsc{NAS} captures a model-agnostic stabilization effect,
rather than exploiting idiosyncrasies of a specific backbone or task, reinforcing its practicality for sequential deployment.

\subsection{Plug-in General Capability Results}
\label{app:plugin_general_capability}

\begin{figure}[H]
  \centering
  \includegraphics[width=\textwidth]{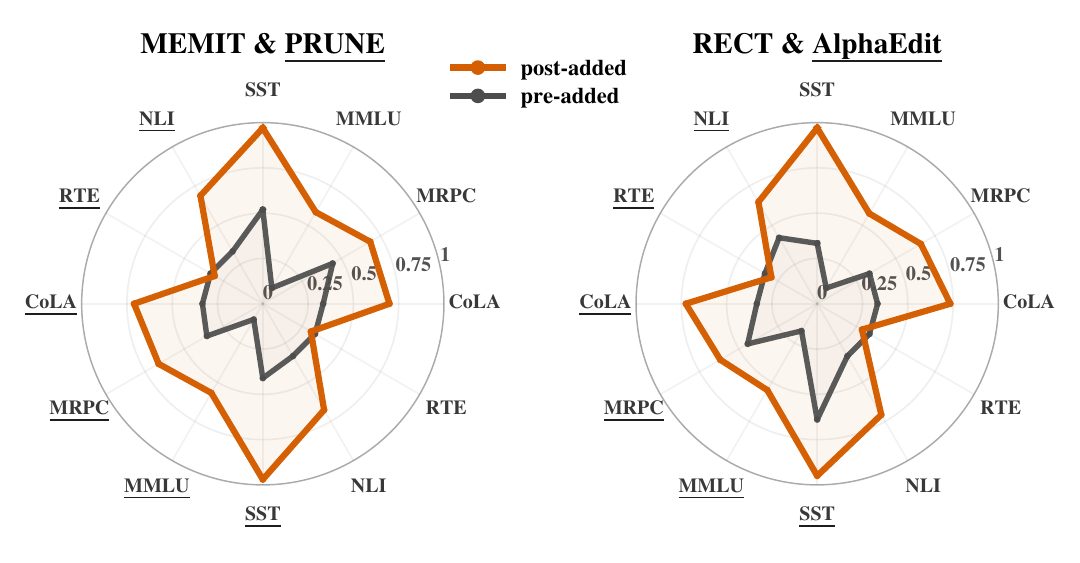}
  \caption{RQ3: \textbf{General capabilities comparison.}
  Radar plots compare pre-added (gray) and post-added (orange) performance on general capabilities (GLUE-style evaluation) for two method pairs: MEMIT \& PRUNE (left) and RECT \& AlphaEdit (right).
  For each radar, the first six axes correspond to the first method in the title, and the underlined six axes correspond to the second method.}
  \label{fig:rq3_radar_cp_general_capabilities}
  \vspace{-3.5mm}
\end{figure}

Figure~\ref{fig:rq3_radar_cp_general_capabilities} compares GLUE-style general capability at the same
reference point used in the plug-in study. For both method pairs (MEMIT \& PRUNE; RECT \& AlphaEdit), the NAS-augmented variants dominate the
base editors on most axes; the only near-exception is RTE, where the gap is comparatively small. Overall, NAS improves
long-horizon editability by mitigating instability, without sacrificing general capabilities.

\clearpage

\subsection{Plug in Experiment on GPT-J.}
\label{app:extraplugin}

\begin{figure}[H]
  \centering
  \includegraphics[width=0.9\textwidth]{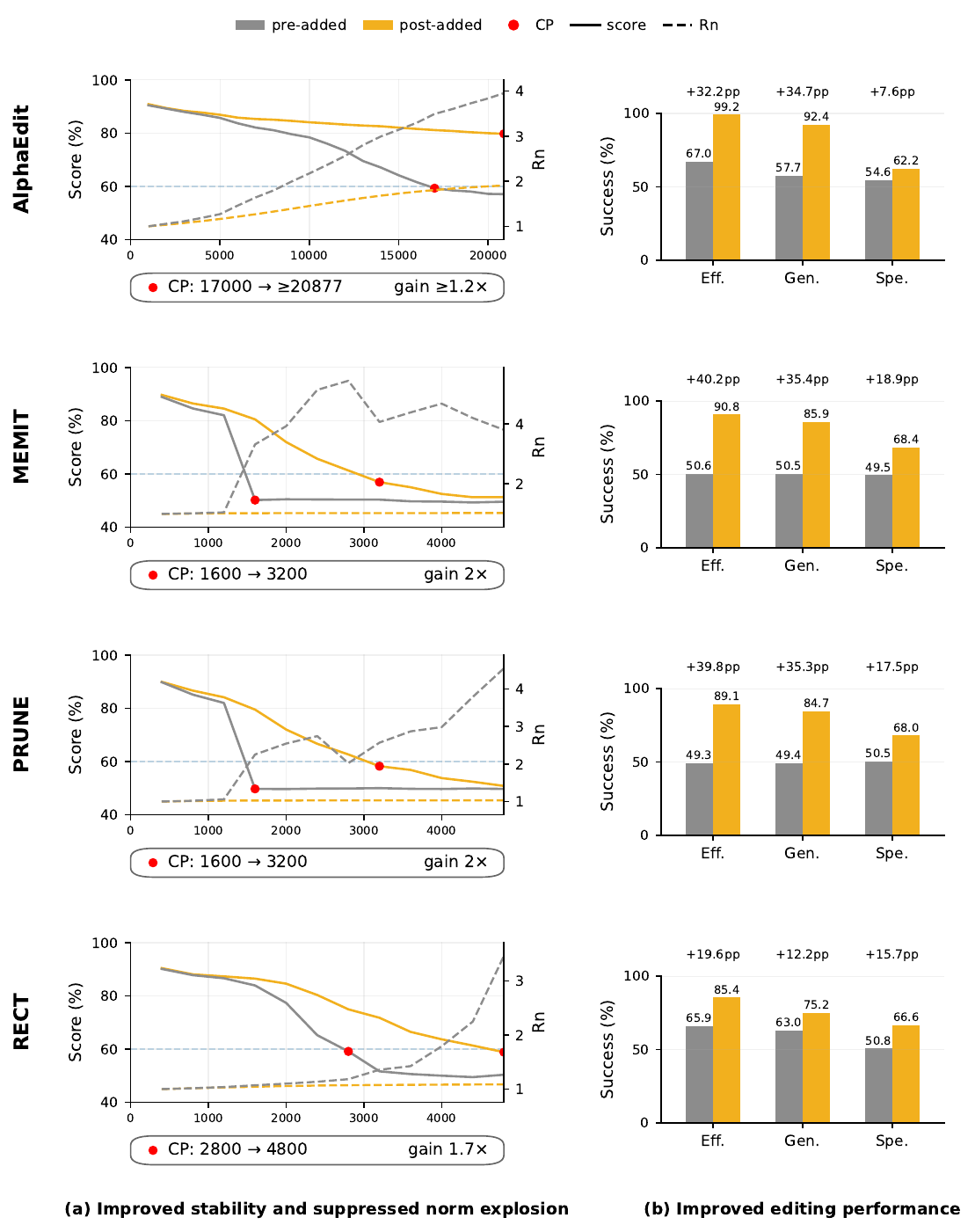}
  \caption{
  \textbf{NAS improves stability and editing performance on GPT-J.}
  \textbf{(a)} Improved stability and suppressed norm explosion: dual-axis trajectories show the editing score (solid) and relative weight norm $R_n$ (dashed); the red dot marks the collapse point (CP, first step with score $\le 60$).
  \textbf{(b)} Improved editing performance: success at the original (w/o NAS) CP, reported on rewrite/para./neighbor (Eff./Gen./Spe.), comparing pre-added vs.\ post-added.
  }
  \label{fig:GPTJ_plug_in}
\end{figure}

We further validate the plug-in behavior of NAS on an additional backbone, GPT-J, using the same experimental setup as
in the main text. We integrate NAS into four representative Locate-and-Edit editors (MEMIT, PRUNE, RECT, and
AlphaEdit) and run long-horizon sequential editing on CounterFact, tracking both the editing score and the relative
weight-norm statistic $R_n$.

Figure~\ref{fig:GPTJ_plug_in} summarizes the results. Across all four editors, attaching NAS reliably suppresses
the growth of $R_n$ and delays the collapse point (CP), indicating that the stabilization effect is not specific to Llama3-8B.
In particular, NAS extends the editable horizon by $2\times$ for MEMIT and PRUNE (CP: 1600 $\rightarrow$ 3200), by
$1.7\times$ for RECT (2800 $\rightarrow$ 4800), and by at least $1.2\times$ for AlphaEdit (17,000 $\rightarrow$ $\ge$\counterfactStreamSize),
where the NAS-augmented run does not collapse within the full CounterFact stream.

In addition to improved stability, NAS also improves editing quality under a shared reference condition.
Evaluated at the original (w/o NAS) CP of each base editor, NAS substantially increases post-edit success on
rewrite/paraphrase/neighborhood (Eff./Gen./Spe.) across methods. The gains are large and consistent, with typical
improvements of roughly $+20$--$+40$ percentage points on efficacy and generalization, and $+8$--$+19$ percentage points on
specificity, mirroring the main-paper findings. Overall, the GPT-J results corroborate that NAS is a portable, plug-and-play
stabilization component that both mitigates instability (via suppressing norm growth) and improves editing performance
across diverse L\&E baselines.

\subsection{Results on Additional Base Models}
\label{app:extrallm}

We further evaluate Locate-and-Edit (L\&E) editors on \textbf{GPT2-XL} under the \textbf{same sequential editing setup as RQ1}.
Table~\ref{tab:seq_edit_gpt2xl_only} reports the long-horizon sequential editing results.
\TabExtraLLM
Consistent with our main findings, several L\&E baselines exhibit severe degradation on ZsRE, collapsing to \textbf{0} post-edit
success for \textsc{MEMIT}/\textsc{PRUNE}/\textsc{RECT}.
\textsc{AlphaEdit} remains substantially more stable, and NAS achieves the best performance
across \emph{all} reported metrics on both CounterFact and ZsRE.
In particular, NAS improves over \textsc{AlphaEdit} by \textbf{+17.8/+15.0/+2.6} points on CounterFact (Eff./Gen./Spe.) and
\textbf{+14.9/+13.4/+2.2} on ZsRE, while also markedly increasing consistency (\textbf{+13.4}).
These results further support the robustness and portability of NAS to additional backbones beyond those in the main text.

\subsection{Additional Baselines under Official Sequential-Editing Configurations}
\label{app:extrabaselines}

\paragraph{Motivation and protocol.}
To complement the main results, we report additional baselines that are not included in the main tables due to
\emph{protocol mismatch}: several non-L\&E editors do not support our \emph{atomic sequential editing} setting
(i.e., updating exactly one fact per step), and are instead designed and tuned for their own official update
granularity and step sizes.
To avoid redefining these methods, we evaluate all additional baselines using their \emph{official} sequential-editing
configurations and learning-rate/step schedules.
MEMOIR does not support our generation-based evaluation, hence Fluency/Consistency entries are marked as NULL.

\paragraph{NAS under two sequential-editing granularities.}
For a controlled comparison against these official settings, we report two variants of our method:
\textsc{NAS} follows the atomic protocol used throughout the paper, whereas \textsc{NAS}$^\dagger$ follows the
batched sequential-editing configuration (updating 500 facts per update step) to match the granularity used
by several additional baselines. Tab.~\ref{tab:seq_edit_vertical_model} summarizes results after 10,000 sequential edits. 

\TabExtraMethod

\paragraph{Summary.}
Across additional baselines and backbones, \textsc{NAS} consistently achieves a favorable efficacy--generalization balance
under sequential editing, while several methods exhibit strong efficacy/locality but weak generalization. We highlight two
representative cases below.

\paragraph{Observations.}
Tab.~\ref{tab:seq_edit_vertical_model} suggests that additional baselines can score highly on efficacy/locality while failing
to generalize. On LLaMA3/CounterFact, GRACE reaches Eff./Spe.=99.08/88.39 (with Flu.=630.42), but its Gen. is 10.20; on
LLaMA3/ZsRE, its Gen. further drops to 1.84. In contrast, \textsc{NAS} yields a more balanced profile on the same backbone:
on CounterFact, \textsc{NAS} achieves Eff./Gen./Spe.=98.85/85.50/64.62 and \textsc{NAS}$^\dagger$ achieves 98.64/88.21/66.13;
on ZsRE, \textsc{NAS} attains Eff./Gen.=92.76/88.71 and \textsc{NAS}$^\dagger$ attains 93.74/89.39.

\paragraph{Example: GPT-J.}
On GPT-J, \textsc{NAS} improves efficacy and generalization simultaneously. On CounterFact, \textsc{NAS} and
\textsc{NAS}$^\dagger$ achieve Eff./Gen.=99.60/93.36 and 99.61/93.92, respectively, compared to GRACE (Gen.=16.40) and
WISE (Gen.=43.02). LyapLock attains high specificity (Spe.=83.05) but with substantially lower Eff./Gen. (47.56/34.60).
On ZsRE, \textsc{NAS}$^\dagger$ achieves Eff./Gen.=97.65/92.78, compared to LyapLock 64.23/58.62.

\paragraph{Transition.}
To further assess the additional editors of the L\&E-paradigm and the effect of enabling NAS as a plug-in, we turn to a longer-horizon evaluation of
WikiBigEdit in the next section.

\subsection{Plug-in Experiments of Additional L\&E methods on WikiBigEdit}
\label{app:wikibigedit}

\begin{figure}[H]
    \centering
    \includegraphics[width=0.8\textwidth]{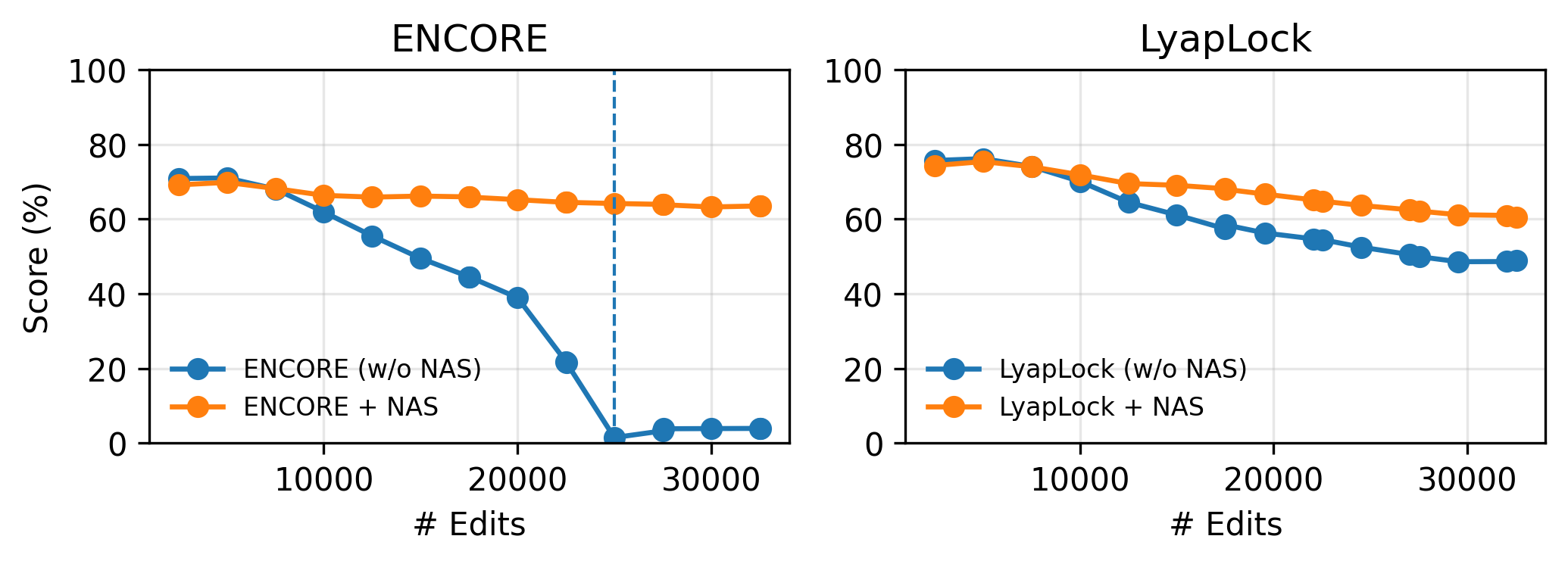}
    \caption{Single-step sequential editing on WikiBigEdit: ENCORE and LyapLock with vs.\ without NAS (Score $=(\mathrm{ES}+\mathrm{GS}+\mathrm{LS})/3$).}
    \label{fig:wikibigedit-s1-plugin}
\end{figure}

\begin{table}[H]
    \centering
    \caption{Single-step sequential editing on WikiBigEdit at 32{,}541 edits (\%; higher is better).}
    \label{tab:wikibigedit-s1-final}
    \begin{tabular}{lcccc}
        \toprule
        Method & ES & GS & LS & Score \\
        \midrule
        ENCORE            & 4.77  & 3.80  & 2.92  & 3.83 \\
        ENCORE + NAS      & 76.83 & 67.75 & \textbf{45.91} & 63.50 \\
        LyapLock          & 62.60 & 55.10 & 29.24 & 48.98 \\
        LyapLock + NAS    & 75.72 & 64.13 & 41.75 & 60.53 \\
        NAS  & \textbf{80.45} & \textbf{68.74} & 43.17 & \textbf{64.12} \\
        \bottomrule
    \end{tabular}
\end{table}

\paragraph{Setup.}
We evaluate long-horizon \emph{atomic sequential editing} on WikiBigEdit, where each edit updates exactly one
fact before proceeding to the next. We periodically evaluate at increasing checkpoints up to 32{,}541 edits.
We report Efficacy (ES), Generalization (GS), and Specificity (LS), and summarize them using
$\mathrm{Score}=(\mathrm{ES}+\mathrm{GS}+\mathrm{LS})/3$.

\paragraph{Additional plug-in results.}
We further enable NAS as a \emph{drop-in} component for existing LE editors by toggling a single additional line of code while keeping other hyperparameters and settings unchanged.
Fig.~\ref{fig:wikibigedit-s1-plugin} compares ENCORE and LyapLock with vs.\ without NAS.
ENCORE without NAS collapses around $\sim$25k edits and degrades to near-zero accuracy thereafter, whereas ENCORE+NAS
remains stable through the full horizon. Concretely, at 32{,}541 edits ENCORE improves from
ES/GS/LS = 4.77/3.80/2.92 (Score 3.83) to 76.83/67.75/45.91 (Score 63.50) with NAS
(Tab.~\ref{tab:wikibigedit-s1-final}).
Similarly, NAS consistently improves LyapLock across checkpoints, yielding ES/GS/LS = 75.72/64.13/41.75
(Score 60.53) at 32{,}541 edits, compared to 62.60/55.10/29.24 (Score 48.98) without NAS.

\paragraph{Reference performance of NAS.}
For reference, we also report results of NAS under the same setting.
At 32{,}541 edits, \textsc{NAS} attains ES/GS/LS = 80.45/68.74/43.17 (Score 64.12)
(Tab.~\ref{tab:wikibigedit-s1-final}).

\paragraph{Takeaway.}
Overall, NAS serves as a simple and effective plug-in that substantially improves long-horizon stability under
single-step sequential editing when combined with prior LE methods.

\subsection{Runtime Overhead of Norm-Anchor Scaling (NAS)}
\label{app:runtime}

We compare the wall-clock runtime with and without NAS under identical editing hyperparameters.
Following our sequential-editing setup on \textbf{Llama3-8B}, we run $n{=}100$ edits for each method and report the
per-edit runtime (mean$\pm$std). We summarize the relative overhead as
$\Delta\% = \frac{t_{\text{NAS}} - t_{\text{base}}}{t_{\text{base}}}\times 100\%$.

\begin{table}[H]
\centering
\caption{
Runtime comparison with/without NAS.
We report mean$\pm$std seconds per edit over $n{=}100$ edits.
$\Delta$\% is computed on the per-edit mean runtime.
}
\label{tab:nas_runtime}
\begin{tabular}{l c c r}
\toprule
\textbf{Method} & \textbf{w/o NAS (s/edit)} & \textbf{w/ NAS (s/edit)} & \textbf{$\Delta$\%} \\
\midrule
MEMIT     & 5.20$\pm$0.82 & 5.20$\pm$0.68 & +0.00 \\
PRUNE     & 5.17$\pm$0.62 & 5.18$\pm$0.68 & +0.19 \\
RECT      & 8.18$\pm$0.84 & 8.08$\pm$0.78 & -1.22 \\
AlphaEdit & 6.75$\pm$0.73 & 6.78$\pm$0.79 & +0.44 \\
\bottomrule
\end{tabular}
\end{table}

Overall, enabling NAS introduces negligible runtime overhead.
Across all tested editors, the per-edit runtime change remains within $\pm 1.3\%$,
and is typically below $0.5\%$.
Given that NAS only adds lightweight vector re-scaling (without extra optimization steps),
the observed differences are well within the natural variance of per-edit runtime
(see the reported std values), supporting that NAS is effectively runtime-free in practice.

\section{Examples of CounterFact and ZsRE dataset}
\label{app:data_examples}

\subsection{CounterFact Examples}
\label{app:counterfact_examples}

\begin{enumerate}
  \item \textbf{Twin-city relation (P190).}
  \begin{itemize}
    \item \textbf{Subject:} Lyon
    \item \textbf{Prompt template:} \texttt{What is the twin city of \{\}? It is}
    \item \textbf{Target (true $\rightarrow$ new):} Beirut $\rightarrow$ Manila
    \item \textbf{Context / paraphrase prompts (examples):}
      \begin{itemize}
        \item \texttt{Lyon is a twin city of}
        \item \texttt{The twin city of Lyon is}
      \end{itemize}
    \item \textbf{Locality prompts (same relation; other subjects, examples):}
      \begin{itemize}
        \item \texttt{What is the twin city of Los Angeles? It is}
        \item \texttt{Athens is a twin city of}
        \item \texttt{The twin city of Beijing is}
      \end{itemize}
    \item \textbf{Neighborhood prompts (subject-related, examples):}
      \begin{itemize}
        \item \texttt{Lyon's twin city is known for}
        \item \texttt{People in Lyon's twin city speak the language of}
      \end{itemize}
  \end{itemize}

  \item \textbf{Mother-tongue relation (P103).}
  \begin{itemize}
    \item \textbf{Subject:} Thomas Joannes Stieltjes
    \item \textbf{Prompt template:} \texttt{The mother tongue of \{\} is}
    \item \textbf{Target (true $\rightarrow$ new):} Dutch $\rightarrow$ English
    \item \textbf{Context / paraphrase prompts (examples):}
      \begin{itemize}
        \item \texttt{Thomas Joannes Stieltjes spoke the language}
        \item \texttt{Thomas Joannes Stieltjes, speaker of}
      \end{itemize}
    \item \textbf{Locality prompts (same relation; other subjects, examples):}
      \begin{itemize}
        \item \texttt{The mother tongue of Rob Birza is}
        \item \texttt{Arend Lijphart is a native speaker of}
        \item \texttt{The native language of Charlie Chaplin is}
      \end{itemize}
  \end{itemize}

  \item \textbf{Citizenship relation (P27).}
  \begin{itemize}
    \item \textbf{Subject:} Mahmoud Fawzi
    \item \textbf{Prompt template:} \texttt{\{\} has a citizenship from}
    \item \textbf{Target (true $\rightarrow$ new):} Egypt $\rightarrow$ Germany
    \item \textbf{Context / paraphrase prompts (examples):}
      \begin{itemize}
        \item \texttt{Mahmoud Fawzi holds a citizenship from}
        \item \texttt{Mahmoud Fawzi, who is a citizen of}
      \end{itemize}
    \item \textbf{Locality prompts (same relation; other subjects, examples):}
      \begin{itemize}
        \item \texttt{Imhotep, who is a citizen of}
        \item \texttt{Marc Forster holds a citizenship from}
        \item \texttt{Katja Ebstein, who is a citizen of}
      \end{itemize}
  \end{itemize}
\end{enumerate}

\subsection{ZsRE Examples}
\label{app:zsre_examples}

\begin{enumerate}
  \item \textbf{Publisher query.}
  \begin{itemize}
    \item \textbf{Subject:} Alien Front Online
    \item \textbf{Prompt:} \texttt{What company published Alien Front Online?}
    \item \textbf{Target (true $\rightarrow$ new):} Sega $\rightarrow$ 2K Games
    \item \textbf{Rephrase prompt:} \texttt{Which company released Alien Front Online?}
    \item \textbf{Locality (Relation\_Specificity, examples):}
      \begin{itemize}
        \item \texttt{The country of origin of Alien Front Online is} $\rightarrow$ Japan
        \item \texttt{Alien Front Online country of origin} $\rightarrow$ Japan
      \end{itemize}
    \item \textbf{Portability (Reasoning, example):}
      \begin{itemize}
        \item \texttt{Who is the parent company of the publisher of Alien Front Online?}
        $\rightarrow$ Take-Two Interactive
      \end{itemize}
  \end{itemize}

  \item \textbf{Programming-language query.}
  \begin{itemize}
    \item \textbf{Subject:} GNOME Chess
    \item \textbf{Prompt:} \texttt{What programming language was used to write GNOME Chess?}
    \item \textbf{Target (true $\rightarrow$ new):} Vala $\rightarrow$ Python
    \item \textbf{Rephrase prompt:} \texttt{How is the programming language for GNOME Chess?}
    \item \textbf{Locality (Relation\_Specificity, examples):}
      \begin{itemize}
        \item \texttt{The platform of GNOME Chess is} $\rightarrow$ Unix-like operating system
        \item \texttt{GNOME Chess platform} $\rightarrow$ Unix-like operating system
      \end{itemize}
    \item \textbf{Portability (Reasoning, example):}
      \begin{itemize}
        \item \texttt{Who created the programming language used to write GNOME Chess?}
        $\rightarrow$ Guido van Rossum
      \end{itemize}
  \end{itemize}

  \item \textbf{Launch-year query.}
  \begin{itemize}
    \item \textbf{Subject:} Old Quebec Street Mall
    \item \textbf{Prompt:} \texttt{When was Old Quebec Street Mall launched?}
    \item \textbf{Target (true $\rightarrow$ new):} 2003 $\rightarrow$ 2002
    \item \textbf{Rephrase prompt:} \texttt{When did Old Quebec Street Mall open?}
    \item \textbf{Locality (Relation\_Specificity, examples):}
      \begin{itemize}
        \item \texttt{The located in the administrative territorial entity of Old Quebec Street Mall is}
        $\rightarrow$ Guelph
        \item \texttt{Old Quebec Street Mall located in the administrative territorial entity}
        $\rightarrow$ Guelph
      \end{itemize}
    \item \textbf{Portability (Reasoning, example):}
      \begin{itemize}
        \item \texttt{What major sporting event took place the same year Old Quebec Street Mall was launched?}
        $\rightarrow$ Salt Lake City Winter Olympics
      \end{itemize}
  \end{itemize}
\end{enumerate}

\section{Broader Impact Discussion}
\label{app:broader_impact}

NAS may improve the reliability of factual updates by reducing unintended degradation during long edit streams.
However, more stable model editing could also make misleading, biased, or harmful factual edits easier to apply at scale.
Responsible use requires auditing edited behavior, tracking edit provenance, and respecting the licenses and access policies of the underlying models and datasets.


\end{document}